\pdfoutput=1

\documentclass[11pt]{article}

\usepackage[]{EMNLP2023}

\usepackage{times}
\usepackage{latexsym}

\usepackage[T1]{fontenc}

\usepackage[utf8]{inputenc}

\usepackage{microtype}

\usepackage{inconsolata}

\usepackage{enumitem}
\usepackage{amstext}
\usepackage{booktabs}
\usepackage{graphicx}
\usepackage{mathtools}
\usepackage{algorithmic}
\usepackage{algorithm}
\usepackage{amsfonts,amssymb}
\usepackage{multirow}
\usepackage{caption}
\usepackage{subcaption}
\usepackage{array}
\usepackage{bm}
\usepackage{makecell}
\usepackage{xcolor}
\usepackage{colortbl}
\usepackage{longtable}

\definecolor{lightgray}{gray}{0.95}
\usepackage{CJKutf8}
\newcommand{\cchar}[1]{\begin{CJK*}{UTF8}{gkai}#1\end{CJK*}}

\title{Sentiment Analysis in the Era of Large Language Models: A Reality Check}

\author{
\textbf{
Wenxuan Zhang$^*$\textsuperscript{\rm 1}~~ 
Yue Deng\thanks{~~Equal contribution. Yue Deng is under the Joint PhD Program between Alibaba and Nanyang Technological University.}~~\textsuperscript{\rm 1,2}~~
Bing Liu\textsuperscript{\rm 3}~~
Sinno Jialin Pan\textsuperscript{\rm 2,4}}~~
Lidong Bing\textsuperscript{\rm 1}~~ \\
\textsuperscript{\rm 1}DAMO Academy, Alibaba Group~~
\textsuperscript{\rm 2}Nanyang Technological University, Singapore\\
\textsuperscript{\rm 3}University of Illinois at Chicago~~
\textsuperscript{\rm 4}The Chinese University of Hong Kong~~
\\
{\tt\{saike.zwx, yue.deng, l.bing\}@alibaba-inc.com} \\
{\tt liub@uic.edu, sinnopan@cuhk.edu.hk}
}

\begin{document}
\maketitle
\begin{abstract}
Sentiment analysis (SA) has been a long-standing research area in natural language processing. It can offer rich insights into human sentiments and opinions and has thus seen considerable interest from both academia and industry. With the advent of large language models (LLMs) such as ChatGPT, there is a great potential for their employment on SA problems. However, the extent to which existing LLMs can be leveraged for different sentiment analysis tasks remains unclear. This paper aims to provide a comprehensive investigation into the capabilities of LLMs in performing various sentiment analysis tasks, from conventional sentiment classification to aspect-based sentiment analysis and multifaceted analysis of subjective texts. We evaluate performance across 13 tasks on 26 datasets and compare the results against small language models (SLMs) trained on domain-specific datasets. Our study reveals that while LLMs demonstrate satisfactory performance in simpler tasks, they lag behind in more complex tasks requiring deeper understanding or structured sentiment information. However, LLMs significantly outperform SLMs in few-shot learning settings, suggesting their potential when annotation resources are limited. We also highlight the limitations of current evaluation practices in assessing LLMs' SA abilities and propose a novel benchmark, \textsc{SentiEval}, for a more comprehensive and realistic evaluation. Data and code during our investigations are available at \url{https://github.com/DAMO-NLP-SG/LLM-Sentiment}.
\end{abstract}

\section{Introduction}
Sentiment analysis\footnote{There are many related terminologies including sentiment analysis, opinion mining, affect analysis, opinion extraction, etc. We collectively refer to them as sentiment analysis in this paper, following the convention in \citet{liubing-sa-book}.} (SA) has been a long established area of research in natural language processing (NLP), which aims to systematically study people's opinions, sentiments, emotions, etc, through computational methods \cite{liubing-sa-book, tac20-sa-survey}. Since its inception \cite{acl02-thumbs, kdd04-sa}, this field has attracted significant interest from both academia and industry due to its wide range of applications, such as product review analysis and gaining insights from social media posts \cite{tweeteval, absa-survey}. Furthermore, achieving a deep understanding of human subjective feeling through sentiment analysis is undoubtedly an important step toward developing artificial general intelligence \cite{sparks-agi}.

In recent years, large language models (LLMs) such as GPT-3 \cite{gpt3-paper}, PaLM \cite{palm-paper}, and GPT-4 \cite{gpt4-report} have demonstrated impressive performance on a wide range of NLP tasks. They can directly perform tasks in zero-shot or few-shot in-context learning manner and achieve strong performance without the need for any supervised training \cite{chatgpt-evaluate, gpt35-evaluate, llm-sa-chatgpt, chatgpt-survey}. Although there have been some initial attempts to apply LLMs to sentiment analysis \cite{www23-llm-sa, llm-sa-chatgpt, llm-sa-xiarui}, these are often limited to some specific tasks within the field and consider different models, datasets, and settings in experiments. As such, the extent to which existing large language models can be leveraged for sentiment analysis remains unclear. 

In this work, we aim to conduct a reality check on the current state of sentiment analysis in the era of large language models. Specifically, we seek to answer the following research questions:
1) \textit{How well do LLMs perform on various sentiment analysis tasks?}
2) \textit{Compared to small specialized models trained on domain-specific datasets, how do large models fare in both zero-shot and few-shot settings?}
3) \textit{Are current SA evaluation practices still suitable to assess models in the era of LLMs?}

To this end, we first conduct a systematic review of various sentiment analysis related tasks, from conventional sentiment classification (SC, classifying the sentiment orientation of a given text) \cite{sst} to aspect-based sentiment analysis (ABSA, analyzing sentiment and opinion information in a more fine-grained aspect-level manner) \cite{absa-survey} and the multifaceted analysis of subjective texts (MAST, focusing on specific sentiment or opinion phenomenon such as hate speech detection and comparative opinion mining) \cite{tweeteval}. In total, we consider 13 sentiment analysis tasks across 26 datasets. These tasks were often studied in isolation due to their unique characteristics in the past. This fragmentation, while necessary in the previous phases, offered a somewhat incomplete understanding of how well models could comprehend human subjective information. With the advent of LLMs, we now have the tools to conduct a more holistic and integrated examination.

For LLMs, we consider both open-source language models such as Flan-T5 \cite{flan-t5-paper} and Flan-UL2 \cite{ul2-paper}, along with GPT-3.5 model series from OpenAI, namely ChatGPT (\texttt{gpt-3.5-turbo}) and \texttt{text-davinci-003} \cite{gpt3-paper, instruct-gpt}.
We also establish comparison baselines using smaller language models\footnote{So far, there is no clear definition of what models can be counted as small or large language models. In this work, we consider model parameters less than 1B as small, and larger than 10B as large for simplified demonstration.} (SLMs) such as T5 \cite{t5-paper}, which allows us to measure the performance of LLMs against these specialized baselines trained with in-domain labeled data. We employ both zero-shot and few-shot settings to evaluate these models across various sentiment analysis tasks, which helps us answer the first two research questions. 

Our investigation yields several insights: Firstly, LLMs already demonstrate satisfactory performance in zero-shot settings for simple SA tasks, such as binary sentiment classification. However, when it comes to more complex tasks, e.g., those requiring a deep understanding of specific sentiment phenomena, or ABSA tasks that necessitate structured sentiment information, LLMs still lag behind SLMs trained with in-domain data. Despite an increased performance can be observed with a larger number of parameters (e.g., from Flan-T5 to ChatGPT), a performance gap remains. Secondly, in the context of few-shot learning, with a limited quantity of annotated data, LLMs consistently outperform SLMs. This suggests that the application of LLMs is advantageous when annotation resources are scarce. Nevertheless, LLMs are constrained by the limited context length for few-shot examples, which needs to be addressed for effective utilization. 

During the investigation, we also identify several limitations of current practice in evaluating a model's SA capability. For example, the evaluations often only involve specific tasks or datasets; and inconsistent prompts are utilized across different studies. While these evaluation practices might have been appropriate in the past, they fall short of accurately assessing LLMs' SA abilities.
To address these issues, we propose a novel benchmark called \textsc{SentiEval}. It breaks the boundary of a wide range of SA tasks, enabling a more comprehensive evaluation of models. It also employs varied task instructions, paired with the corresponding text, alleviating the sensitivities associated with prompt design during the evaluation of different LLMs. Furthermore, by framing these tasks as natural language instructions, we create a more realistic evaluation environment akin to a real-world practical use case.

\section{Background}
\subsection{Sentiment Analysis}
Sentiment analysis has received lots of attention since its early appearance \cite{acl02-thumbs, emnlp02-yu-sa, kdd04-sa} and remained an active research area in the field of NLP nowadays \cite{liubing-sa-book, tac20-sa-survey, aireview20-sa-dl-survey}. This enduring interest mainly stems from two aspects. Firstly, the ability to comprehend the subjective sentiments and opinions within textual data is a critical step toward achieving human-level intelligence \cite{sparks-agi}. For example, understanding human emotions, recognizing their dynamic changes, and providing emotional responses are key elements in creating human-like chatbots \cite{acl19-empathetic-dialog, acl21-emotional-dialog}. Secondly, the practical applications of sentiment analysis span a broad spectrum, especially with the explosive growth of user-generated content in the past decades. SA has found extensive applications such as analyzing customer reviews \cite{emnlp20-multilingual-amazon-review, absa-survey}, monitoring social media opinions \cite{sa-survey-scoial-media, tweeteval}, etc.

Given its importance, sentiment analysis comprises a broad spectrum of tasks for understanding and analyzing human sentiment, emotion, and subjective feeling in the text. One of the earliest and most fundamental tasks is the sentiment classification \cite{acl02-thumbs}, which aims at determining the overall sentiment polarity of a given text, typically in a binary (positive, negative) or multi-class (positive, neutral, negative) format \cite{emnlp20-multilingual-amazon-review}. In recent years, with the more powerful deep learning models, two directions have appeared which either go ``deep'' or go ``wide''. The deep direction moves towards more granular tasks, namely aspect-based sentiment analysis (ABSA). ABSA aims to extract detailed sentiment information about specific aspects or features of an opinion target \cite{absa-survey}. Another direction extends SA to the multifaceted analysis of subjective texts (MAST), which encompasses various specialized tasks focusing on specific sentiment or opinion phenomena  \cite{liubing-sa-book}. For example, hate speech detection aims to identify aggressive or derogatory sentiments targeted toward specific groups \cite{hate-survey}. Other tasks include irony detection \cite{irony-survey}, comparative opinion mining \cite{comparative-survey}, emotion detection \cite{emotion-survey} etc, each addressing different dimensions of sentiment in text. All these tasks collectively contribute to a holistic understanding of sentiment in language and demonstrate the wide range of tasks falling under the umbrella of sentiment analysis.

\subsection{Large Language Models}
Recently, there has been a remarkable advancement in the development of large language models (LLMs), such as GPT-3 \cite{gpt3-paper}, PaLM \cite{palm-paper}, Flan-UL2 \cite{ul2-paper}, LLaMA \cite{llama} and ChatGPT.
These LLMs conduct pre-training on large amounts of text data and employ various training techniques, including instruction tuning \cite{instruction-tuning}, reinforcement learning from human feedback (RLHF) \cite{RLHF} and etc.
As a result, LLMs demonstrate impressive capabilities in zero-shot or few-shot learning settings, thereby shifting the focus of NLP from the fine-tuning paradigm toward the prompting paradigm.

There are some initial attempts on evaluating LLMs for SA tasks. \citet{llm-sa-chatgpt} observe that the zero-shot performance of LLMs is comparable to fine-tuned BERT model. 
In addition, \citet{llm-sa-xiarui} conduct a preliminary study with ChatGPT for some SA tasks, specifically investigating its ability to handle polarity shifts, open-domain scenarios, and sentiment inference problems.
Moreover, \citet{www23-llm-sa} explore the fine-tuning of a small student model with an LLM to generate weak labels, and the final model performs on par with existing supervised models.
Despite those existing efforts, their scope is often limited to specific tasks and involves different datasets and experimental designs. The true capacity of LLMs for sentiment analysis remains unclear, and we aim to conduct a reality check in this paper.

\section{Investigated Tasks and Datasets}
We conduct an extensive survey of a wide range of SA tasks and categorize different tasks into three types: sentiment classification (SC), aspect-based sentiment analysis (ABSA), and multifaceted analysis of subjective texts (MAST). 
We describe investigated tasks of each type, along with the datasets and evaluation metrics.
To ensure balance across various tasks and datasets, we limit our evaluation by sampling a maximum of 500 examples from the test set of each dataset.
Detailed statistics on each task and dataset are summarized in Table \ref{tab:statistics}.

\begin{table*}[!t]
    \centering
    \resizebox{0.88\linewidth}{!}{
    \begin{tabular}{cl|rrr|c|cc}
    \toprule
    Task & Dataset & train & dev & test & sampled test & $\text{class}^*$ & metric \\
    \midrule
    \multicolumn{8}{c}{\textit{Sentiment Classification (SC)}} \\
    \midrule
    \rowcolor{lightgray}
     & IMDb & 22,500 & 2,500 & 25,000 & 500 & 2 & accuracy \\
    \rowcolor{lightgray}
    & Yelp-2 & 504,000 & 56,000 & 38,000 & 500 & 2 & accuracy\\
    \rowcolor{lightgray}
    \multirow{-3}{*}{\parbox{1.9cm}{Document-Level}} & Yelp-5 & 585,000 & 65,000 & 50,000 & 500 & 5 & accuracy\\
     & MR & 8,530 & 1,066 & 1,066 & 500 & 2 & accuracy \\
    & SST-2 & 6,920 & 872 & 1,821 & 500 & 2 & accuracy\\
    & Twitter & 45,615 & 2,000 & 12,284 & 500 & 3 & accuracy \\ 
    \multirow{-4}{*}{\parbox{1.9cm}{Sentence-Level}} & SST-5 & 8,544 & 1,101 & 2,210 & 500 & 5 & accuracy\\
    \rowcolor{lightgray}
     & lap14 & 2,282 & 283 & 632 & 500 & 3 & accuracy \\
    \rowcolor{lightgray}
    \multirow{-2}{*}{\parbox{1.9cm}{Aspect-Level}} & rest14 & 3,608 & 454 & 1,119 & 500 & 3 & accuracy\\
    \midrule
    \multicolumn{8}{c}{\textit{Aspect-based Sentiment Analysis (ABSA)}} \\
    \midrule
     & Rest14 & 2,736 & 304 & 800 & 500 & 3 & micro\_f1 \\
     & Rest15 & 1,183 & 130 & 685 & 500 & 3 & micro\_f1 \\
     & Rest16 & 1,799 & 200 & 676 & 500 & 3 & micro\_f1 \\
    \multirow{-4}{*}{\parbox{1.2cm}{UABSA}} & Laptop14 & 2,741 & 304 & 800 & 500 & 3 & micro\_f1\\
    \rowcolor{lightgray}
     & Rest14 & 1,266 & 310 & 492 & 492 & 3 & micro\_f1 \\
    \rowcolor{lightgray}
     & Rest15 & 605 & 148 & 322 & 322 & 3 & micro\_f1 \\
    \rowcolor{lightgray}
     & Rest16 & 857 & 210 & 326 & 326 & 3 & micro\_f1 \\
    \rowcolor{lightgray}
    \multirow{-4}{*}{\parbox{1.2cm}{ASTE}} & Laptop14 & 906 & 219 & 328 & 328 & 3 & micro\_f1\\
     & Rest15 & 834 & 209 & 537 & 500 & 13 & micro\_f1 \\
    \multirow{-2}{*}{\parbox{1.2cm}{ASQP}} & Rest16 & 1,264 & 316 & 544 & 500 & 13 & micro\_f1\\
    \midrule
    \multicolumn{8}{c}{\textit{Multifaceted Analysis of Subjective Text (MAST)}} \\
    \midrule
    \rowcolor{lightgray}
    Implicit & Lap+Res & 1,746 & NA & 442 & 442 & 3 & accuracy \\
    Hate & HatEval & 9,000 & 1,000 & 2,970 & 500 & 2 & macro\_f1 \\
    \rowcolor{lightgray}
    Irony & Irony18 & 2,862 & 955 & 784 & 500 & 2 & f1(irony) \\
    Offensive & OffensEval & 11,916 & 1,324 & 860 & 500 & 2 & macro\_f1 \\
    \rowcolor{lightgray}
    Stance & Stance16 & 2,620 & 294 & 1,249 & 500 & 3 & $\text{macro}\_\text{f1}^\dagger$\\
    Comparative & CS19 & 1,094 & 157 & 314 & 314 & 2 & accuracy \\
    \rowcolor{lightgray}
    Emotion & Emotion20 & 3,257 & 374 & 1,421 & 500 & 4 & macro\_f1 \\
    \bottomrule
    \end{tabular}}
    \caption{Investigated tasks and dataset statistics. $*$ represents the number of sentiment classes among each task, except for the two datasets of ASQP, which represent the number of aspect categories. $\dagger$ denotes the macro\_f1 score without none class.}
    \label{tab:statistics}
\end{table*}

\subsection{Sentiment Classification}
\label{sec:sc}
Sentiment classification (SC) aims at assigning predefined sentiment classes (e.g., positive, negative, or neutral) to given texts \cite{liubing-sa-book}. 
It serves as a fundamental measure of sentiment orientation and is commonly used to analyze customer reviews, social media posts and etc. 
It can involve a varying number of sentiment classes, ranging from binary classification, where sentiments are categorized as either positive or negative, to more nuanced five-class classification, which grades sentiments on a scale from very negative to very positive.
There are also different levels of granularity at which sentiment can be analyzed, including document-level, sentence-level, and aspect-level SC.

\paragraph{Document-Level}
Sentiment classification at the document level aims to determine the overall sentiment expressed in a text corpus, providing a high-level understanding of the expressed sentiment orientation.
We evaluate on three widely used datasets, including IMDb \cite{imdb}, Yelp-2, and Yelp-5 \cite{yelp}.
The IMDb dataset contains movie reviews, whereas the Yelp-2 dataset includes customer reviews for businesses. Reviews of both datasets are labeled as either \textit{positive} or \textit{negative}. However, the Yelp-5 dataset offers a more fine-grained sentiment classification by introducing three additional sentiment classes: \textit{very positive}, \textit{very negative}, and \textit{neutral}.
We employ accuracy as the evaluation metric.

\paragraph{Sentence-Level}
Sentence-level classification allows for sentiment analysis on a sentence-by-sentence basis. It is particularly useful in analyzing social media posts, customer feedback, or any text where sentiments may change rapidly from sentence to sentence.
We select multiple datasets for evaluation, including MR \cite{mr}, SST2, SST5 \cite{sst}, and Twitter \cite{twitter-sc}.  
The MR, SST2, and SST5 datasets contain movie reviews, whereas the Twitter dataset consists of social media posts. 
While the SST2 and MR datasets use binary sentiment labels, Twitter's sentiment analysis introduces an additional \textit{neutral} class.
In addition, SST5 provides a wider range of labels including \textit{very positive}, \textit{positive}, \textit{neutral}, \textit{negative}, and \textit{very negative} sentiments. 
To evaluate the performance on these datasets, we use accuracy as a metric.

\paragraph{Aspect-Level}
Since sentiment expressed towards different targets might be different even within a single sentence, aspect sentiment classification dives even deeper into the analysis by focusing on identifying sentiment towards specific aspects or entities mentioned. This level of analysis is particularly valuable when the sentiment towards different aspects or entities needs to be assessed individually. 
There are two widely used datasets including Lap14 and Rest14. 
These datasets were introduced in the SemEval ABSA challenge 2014 \cite{semeval14} and consist of laptop and restaurant reviews, respectively. 
The goal is to determine the sentiment towards a specific aspect mentioned in a review sentence, classifying it as either \textit{positive}, \textit{negative}, or \textit{neutral}.
Performance assessment is based on the metric of accuracy.

\subsection{Aspect-based Sentiment Analysis}
\label{sec:absa}
Aspect-based sentiment analysis (ABSA) refers to the process of analyzing people's sentiments at a more fine-grained aspect level. 
It encompasses the analysis of various sentiment elements, such as aspects, opinions, and sentiment polarities \cite{absa-survey}.
ABSA has gained significant attention in recent years, resulting in the emergence of a wide range of tasks. We focus on three compound ABSA tasks here for investigation, which aim to jointly extract multiple sentiment elements. 

\paragraph{Unified Aspect-based Sentiment Analysis (UABSA)}
UABSA is the task of extracting both the aspect and its corresponding sentiment polarity simultaneously.
We evaluate UABSA on four datasets originally from SemEval-2014 \cite{semeval14}, SemEval-2015 \cite{semeval15}, and SemEval-2016 \cite{semeval16} shared tasks, which consist of reviews from Laptops and Restaurants domains.
Following previous studies, we use Micro-F1 score as the metric for evaluation. A predicted pair would be counted as correct only if both the aspect term and sentiment polarity match exactly with the gold labels.

\paragraph{Aspect Sentiment Triplet Extraction (ASTE)}
The ASTE task further extracts the opinion terms on the basis of the UABSA task, which provides an explanation for the predicted sentiment on certain aspects. Therefore, the final target of ASTE is to extract the (aspect, opinion, sentiment) triplet for a given text. 
The datasets we utilized were introduced by \citet{xu-etal-2020-aste}, which were built upon four UABSA datasets. Likewise, we employ the Micro-F1 metric and consider an exact match prediction of each triplet as correct.

\paragraph{Aspect Sentiment Quadruple Prediction (ASQP)}
ASQP task was introduced to provide a complete aspect-level sentiment structure, namely (category, aspect, opinion, sentiment) quadruple \cite{asqp, acl20-acos}. By introducing an additional aspect category element, it can still provide useful information when the aspect term is not explicitly mentioned.
Our study utilizes two restaurant datasets from \citet{asqp}.
We adopt the same evaluation metric and standardization with UABSA and ASTE, using Micro-F1 score as the evaluation metric.

\subsection{Multifaceted Analysis of Subjective Text}
\label{sec:ac}
Multifaceted analysis of subjective text (MAST) are tasks that involve different aspects of human subjective feeling reflected in the text \cite{liubing-sa-book, tac20-sa-survey}. These tasks expand SA beyond merely identifying positive or negative feelings but focus on recognizing and understanding a broader range of human emotional states.

\paragraph{Implicit Sentiment Analysis}
Implicit sentiment analysis focuses on identifying the sentiment expressed indirectly or implicitly in text. 
It requires uncovering sentiments that are conveyed through subtle cues, such as contextual clues, tone, or linguistic patterns.
\citet{implicit} divided the Laptop and Restaurant reviews from SemEval 2014 \cite{semeval14} into two parts: implicit and explicit. For our analysis, we only utilized the implicit dataset and merged the data from both domains into a single dataset. To evaluate the performance, we employed accuracy as the metric.

\paragraph{Hate Speech Detection}
Hate speech detection refers to the process of identifying content that promotes discrimination, hostility, or violence against individuals or groups based on attributes such as race, religion, ethnicity, gender, sexual orientation, or other protected characteristics \cite{hate-survey}.
For our analysis, we utilize the dataset from the SemEval2019 HatEval challenge \cite{hate}. 
This dataset focuses on predicting whether a tweet exhibits hateful content towards two specific target communities: immigrants and women.
We calculate the macro-averaged F1 score across the two binary classes: \textit{hate} and \textit{non-hate}.

\paragraph{Irony Detection}
Irony is a rhetorical device where the intended meaning of a statement is different or opposite to its literal interpretation.
Irony detection aims to recognize and understand instances of irony in the text \cite{irony-survey}.
We choose the Subtask 3A dataset of the SemEval2018 Irony Detection challenge \cite{irony} (referred to as ``Irony18'').
The goal is to determine whether a tweet contains ironic intent or not. For evaluation, we follow the convention to specifically consider the F1 score for the \textit{irony} class, while ignoring \textit{non-irony} F1 score.

\paragraph{Offensive Language Identification}
Offensive language identification involves identifying and flagging text that contains offensive or inappropriate content, including profanity, vulgarities, obscenities, or derogatory remarks \cite{offensive-survey}.
Different from hate speech, offensive language does not necessarily target a specific individual or group. For example, profanity expressions can be considered offensive language even when not directed at anyone in particular.
We use the SemEval2019 OffensEval dataset \cite{offensive}.
It involves classifying each given text as either \textit{offensive} or \textit{non-offensive}.
We adopt macro-averaged F1 score as the metric.

\paragraph{Stance Detection}
Stance detection refers to determining the perspective or stance expressed in a given text towards a particular topic or entity. It helps identify whether the text expresses \textit{favor}, \textit{against}, or \textit{none} opinion towards a subject \cite{stance-survey}.
We utilize the SemEval2016 shared task on Detection Stance in Tweets \cite{stance}, and refer to it as ``Stance16''.
It provides data in five domains (i.e., targets): abortion, atheism, climate change, feminism, and Hillary Clinton.
In order to facilitate evaluation, we aggregate these domains into a single dataset.
When evaluating the results, we only consider macro-averaged of F1 of \textit{favor} and \textit{against} classes, and ignore \textit{none} class, following previous studies.

\paragraph{Comparative Opinion Mining}
Comparative opinion mining is the task of analyzing opinions and sentiments expressed in a comparative context \cite{comparative-survey}. 
It involves  comparing different aspects of a product, service, or any other subject  to determine preferences or relative opinions.  
In our study, we take the CS19 dataset \cite{comparative-CS19}, which provides annotated comparative sentences in the field of computer science. 
These sentences involve comparisons between various targets such as programming languages, database products, and technology standards. 
The opinions expressed in the dataset are categorized as either \textit{better} or \textit{worse}.
To evaluate the performance, we employ accuracy as the metric.

\paragraph{Emotion Recognition}
Emotion recognition involves the identification and understanding of emotions expressed in text \cite{emotion-survey}.
It focuses on detecting and categorizing different emotional states. 
We use the dataset provided by the TweetEval benchmark \cite{tweeteval}, which we refer to it as ``Emotion20''.
It transforms the SemEval2018 Affects in Tweets dataset \cite{emotion-semeval} from multi-class classification into a multi-label dataset, by keeping only the tweets labeled with a single emotion. It selects the most common four emotions, namely \textit{anger}, \textit{joy}, \textit{sadness}, and \textit{optimism}.
For evaluation, we utilize the macro-averaged F1 score, which considers the overall performance across all classes.

\begin{figure*}[t]
    \centering
    \includegraphics[width=\linewidth]{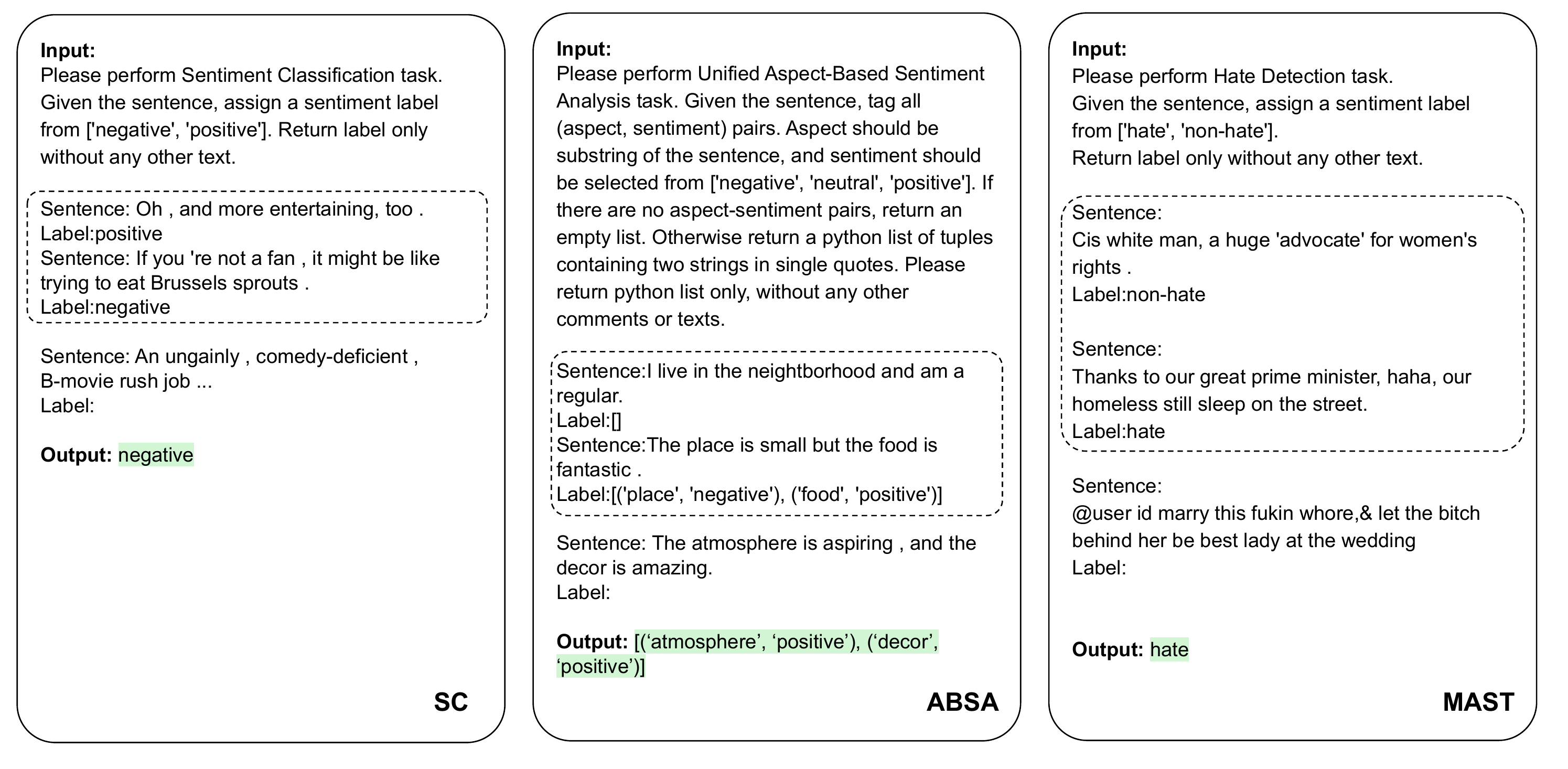}
    \caption{Prompt examples for SC, ABSA, and MAST respectively. The text inside the dashed box are demonstrations of the few-shot setting, and would be removed under the zero-shot setting. }
    \label{fig:prompt}
    \vspace{-0.3cm}
\end{figure*}

\section{Evaluations}
\subsection{Models and Baselines} \label{sec:llms}
\paragraph{Large Language Models (LLMs)} For large language models, we mainly investigate their performance when directly conducting inference on the downstream SA tasks without specific training. We adopt two models from the Flan model family since they are open-sourced and showed strong zero-shot and few-shot performance, namely \texttt{Flan-T5} (XXL version, 13B) \cite{flan-t5-paper} and \texttt{Flan-UL2} (20B) \cite{ul2-paper}. 
We use their checkpoints hosted on Huggingface for the inference. We also take two models from OpenAI, including \texttt{ChatGPT} (\texttt{gpt-3.5-turbo}\footnote{May 12 version of ChatGPT is used for the experiments. It should be noted that future updates might potentially impact the outcomes presented in this paper.}) and the text-davinci-003 model (\texttt{text-003}, 175B) of the GPT-3.5 family. All the temperatures of these models are set to zero for deterministic predictions.

\paragraph{Small Language Models (SLMs)} For small language models, we take T5 (large version, 770M) \cite{t5-paper}, which shows great performance in tackling multiple tasks in the unified text-to-text format. We train the T5 model with domain-specific data on each dataset, with either the full training set (statistics detailed in Table~\ref{tab:statistics}) or sampled data in the few-shot setting.
We use the Adam optimizer with a learning rate of 1e-4, and a fixed batch size of 4 for all tasks.
Regarding training epochs, we select 3 for the full training setting and 100 for the few-shot training setting. We conduct three runs with different random seeds for SLMs and report the average results for more stable comparisons.

\subsection{Prompting Strategy} \label{sec:prompt-strategy}
LLMs may produce very different responses even when the prompts are semantically similar \cite{nips21-prompt-varaince, acl22-prompt-varaince}. 
Furthermore, the preference for prompts varies from one LLM to another.
Therefore, we aim to provide relatively consistent prompts for all datasets across different models in this study, rather than specific designs, in order to evaluate the general performance of LLMs.
Our goal is to design prompts that are simple, clear, and straightforward.

For zero-shot learning, we include only essential components in the prompt, namely the task name, task definition, and output format.
The task name serves the purpose of identifying and specifying the task. 
The task definition is constructed based on each task's definition and annotation guidelines, and also incorporates the label space as a set of options for the model to output its response.
The output format defines the expected structure of the output, enabling us to decode the model's responses into our desired format.
For few-shot learning, an additional ``demonstration'' part is added.
This includes $k$ examples for each class, each accompanied by their respective gold labels in the desired format.
We provide illustrative examples for each task type in Figure \ref{fig:prompt}.
For more detailed information and examples, please refer to Appendix \ref{prompt_summary}.

\subsection{Zero-shot Results}
We summarize the zero-shot performance in Table~\ref{tab:zero-shot}. Two baselines are further included for better comparisons: \texttt{random} assigns a random label to each sample, and \texttt{majority} takes the most common label from the training set's label distribution as the prediction. For LLMs, we utilize them directly to infer the results on the test sets of each dataset. For SLMs, we employ the complete training set to train the model before proceeding to conduct inference on the same test set. 
The following observations can be made.

\begin{table*}[!t]
    \centering
    \resizebox{0.95\linewidth}{!}{
    \begin{tabular}{cl|cc|cccc|c}
    \toprule
    \multirow{3}{*}{Task} & \multirow{3}{*}{Dataset} & \multicolumn{2}{c}{Baseline} & \multicolumn{4}{c}{LLM} & SLM \\
    \cmidrule(lr){3-4} \cmidrule(lr){5-8} \cmidrule(lr){9-9} 
    & & \texttt{random} & \texttt{majority} & \texttt{Flan-T5} & \texttt{Flan-UL2} & \texttt{text-003} & \texttt{ChatGPT} & \texttt{T5}$_{large}$ \\
    & & - & - & (11B) & (20B) & (175B) & (NA) & (770M) \\
    \midrule
    \multicolumn{9}{c}{\textit{Sentiment Classification (SC)}} \\
    \midrule
    \rowcolor{lightgray}
    & IMDb & 52.40 & 46.80 & 86.60 & 97.40 & 90.60 & 94.20 & 93.93 \\
    \rowcolor{lightgray}
    & Yelp-2 & 52.80 & 48.00 & 92.20 & 98.20 & 93.20 & 97.80 & 96.33 \\
    \rowcolor{lightgray}
    \multirow{-3}{*}{\parbox{1.2cm}{Document-Level}} & Yelp-5 & 19.80 & 18.60 & 34.60 & 51.60 & 48.60 & 52.40 & 65.60\\
    \multirow{4}{*}{\parbox{1.2cm}{Sentence-Level}} & MR & 47.40 & 49.60 & 66.00 & 92.20 & 86.80 & 89.20 & 90.00 \\
    & SST2 & 49.20 & 48.60 & 72.00 & 96.40 & 92.80 & 93.60 & 93.20 \\
    & Twitter & 34.20 & 45.40 & 43.60 & 47.40 & 59.40 & 69.40 & 67.73 \\
    & SST5 & 21.40 & 22.20 & 15.00 & 57.00 & 45.20 & 48.00 & 56.80 \\
    \rowcolor{lightgray}
    & Lap14 & 34.80 & 53.80 & 69.00 & 73.20 & 74.60 & 76.80 & 78.60 \\
    \rowcolor{lightgray}
    \multirow{-2}{*}{\parbox{1.2cm}{Aspect-Level}} & Rest14 & 34.00 & 65.60 & 80.80 & 82.40 & 80.00 & 82.80 & 83.67 \\
    \multicolumn{2}{c|}{\textbf{Average}} & 38.44 & 44.29 & 62.20 & 77.31 & 74.58 & 78.24 & 80.65 \\
    \midrule
    \multicolumn{9}{c}{\textit{Aspect-Based Sentiment Analysis (ABSA)}} \\
    \midrule
    \rowcolor{lightgray}
     & Rest14 & NA & NA & 0.00 & 0.00 & 47.56 & 54.46 & 75.31 \\
    \rowcolor{lightgray}
    & Rest15 & NA & NA & 0.00 & 0.00 & 35.63 & 40.03 & 65.46 \\
    \rowcolor{lightgray}
    & Rest16 & NA & NA & 0.00 & 0.00 & 40.85 & 75.80 & 73.23 \\
    \rowcolor{lightgray}
    \multirow{-4}{*}{\parbox{1.2cm}{UABSA}}& Laptop14 & NA & NA & 0.00 & 0.00 & 28.63 & 33.14 & 62.35 \\
    \multirow{4}{*}{\parbox{1.2cm}{ASTE}} & Rest14 & NA & NA & 0.00 & 0.00 & 41.43 & 40.04 & 65.20 \\
     & Rest15 & NA & NA & 0.00 & 0.00 & 37.53 & 33.51 & 57.78 \\
     & Rest16 & NA & NA & 0.00 & 0.00 & 41.03 & 42.18 & 65.94 \\
     & Laptop14 & NA & NA & 0.00 & 0.00 & 27.05 & 27.30 & 53.69 \\
    \rowcolor{lightgray}
     & Rest15 & NA & NA & 0.00 & 0.00 & 13.73 & 10.46 & 41.08 \\
    \rowcolor{lightgray}
     \multirow{-2}{*}{\parbox{1.2cm}{ASQP}}  & Rest15 & NA & NA & 0.00 & 0.00 & 18.18 & 14.02 & 50.58 \\
    \multicolumn{2}{c|}{\textbf{Average}} & NA & NA & 0.00 & 0.00 & 33.16 & 37.09 & 61.06 \\
    \midrule
    \multicolumn{9}{c}{\textit{Multifaceted Analysis of Subjective Text (MAST)}} \\
    \midrule
    \rowcolor{lightgray}
    Implicit & Lap+Res & 35.75 & 56.11 & 33.03 & 42.53 & 45.25 & 54.98 & 67.12 \\
    Hate & HatEval & 48.00 & 36.31 & 56.09 & 70.80 & 67.79 & 50.92 & 46.94 \\
    \rowcolor{lightgray}
    Irony & Irony18 & 50.96 & 58.96 & 27.31 & 73.84 & 76.61 & 68.66 & 79.44 \\
    Offensive & OffensEval & 46.67 & 41.86 & 32.78 & 74.44 & 73.31 & 64.88 & 80.76 \\
    \rowcolor{lightgray}
    Stance & Stance16 & 33.94 & 35.82 & 20.74 & 61.10 & 39.96 & 50.25 & 67.33 \\ 
    Comparative & CS19 & 49.36 & 73.89 & 54.46 & 85.67 & 74.52 & 75.80 & 89.49 \\
    \rowcolor{lightgray}
    Emotion & Emotion20 & 22.87 & 13.92 & 44.34 & 69.92 & 70.51 & 72.80 & 80.35 \\
    \multicolumn{2}{c|}{\textbf{Average}} & 41.08 & 45.27 & 38.39 & 68.33 & 63.99 & 62.61 & 73.05 \\ 
     
    \bottomrule
    \end{tabular}}
    \caption{Zero-shot performance of various sentiment analysis tasks. Similar to GLUE \cite{glue}, "Average" rows show the average of all dataset-specific metrics.
    }
    \label{tab:zero-shot}
\end{table*}

\paragraph{LLMs such as ChatGPT demonstrate strong zero-shot performance in simple SA tasks.} As can be observed in the top and bottom parts of Table~\ref{tab:zero-shot}, LLMs have demonstrated a strong ability to tackle simple SC tasks such as binary sentiment classification and MAST tasks without any prior training. For example, \texttt{ChatGPT} achieves comparable results to the \texttt{T5} model, which has been specifically fine-tuned with the full training set for each dataset. On average, \texttt{ChatGPT}'s performance reaches 97\% of the \texttt{T5}'s prediction on SC tasks, and 83\% on MAST tasks, respectively.
This suggests a superior sentiment analysis ability already inherent in these models. However, we can notice that for more complicated tasks, it still lags behind the fine-tuned models, e.g., 52.4 v.s. 65.6 accuracy scores on Yelp-5 datasets which is a fine-grained five-class SC task, and 72.80 v.s. 80.35 accuracy scores on the comparative opinion mining task.

\paragraph{Larger models do not necessarily lead to better performance.} One observation made from analyzing the performance change among those LLMs is that larger models, with a greater number of parameters, tend to outperform the smaller ones, e.g., comparing the performance between \texttt{Flan-T5} and \texttt{text-003}. However, this does not necessarily mean that scaling up the model size always leads to better results. For instance, \texttt{Flan-UL2}, despite not being the largest model, is able to achieve comparable, and in some cases, superior performance to larger models like \texttt{text-003} across multiple tasks, possibly due to the advantage of both reasonable model size and large-scale instruction tuning.

\paragraph{LLMs struggle with extracting fine-grained structured sentiment and opinion information.} While LLMs have shown proficiency in many SA tasks, they fall short when it comes to extracting structured and fine-grained sentiment and opinion information. For instance, \texttt{Flan-T5} and \texttt{Flan-UL2} were unable to achieve any notable performance on any ABSA tasks across all datasets, as can be noted from the middle part of Table~\ref{tab:zero-shot}. \texttt{text-003} and \texttt{ChatGPT} provide better results but were still significantly outperformed by fine-tuned smaller language models. For example, \texttt{text-003} reaches only around 54\% of the performance of a fine-tuned \texttt{T5} model, though being more than 200 times larger.

\paragraph{RLHF may lead to unexpected phenomena.} An unexpected and interesting observation is that \texttt{ChatGPT} performs poorly in detecting hate speech, irony, and offensive language. Even compared to \texttt{text-003}, which archives similar performance on many other tasks, \texttt{ChatGPT} still performs much poorer on these three tasks. One possible explanation for this could be an "over-alignment" with human preference during the RLHF process of training \texttt{ChatGPT}~\cite{RLHF}. This phenomenon suggests that these models, in their quest to mimic human-like conversation and sentiment, may inadvertently adopt human biases or become over-sensitive to certain types of negative or offensive speech patterns. This finding emphasizes the need for further research and improvements in these areas.

\subsection{Analysis of Sensitivity on Prompt Design} \label{sec:sensitity-prompt}
\begin{figure}[t]
    \centering
    \includegraphics[width=\linewidth]{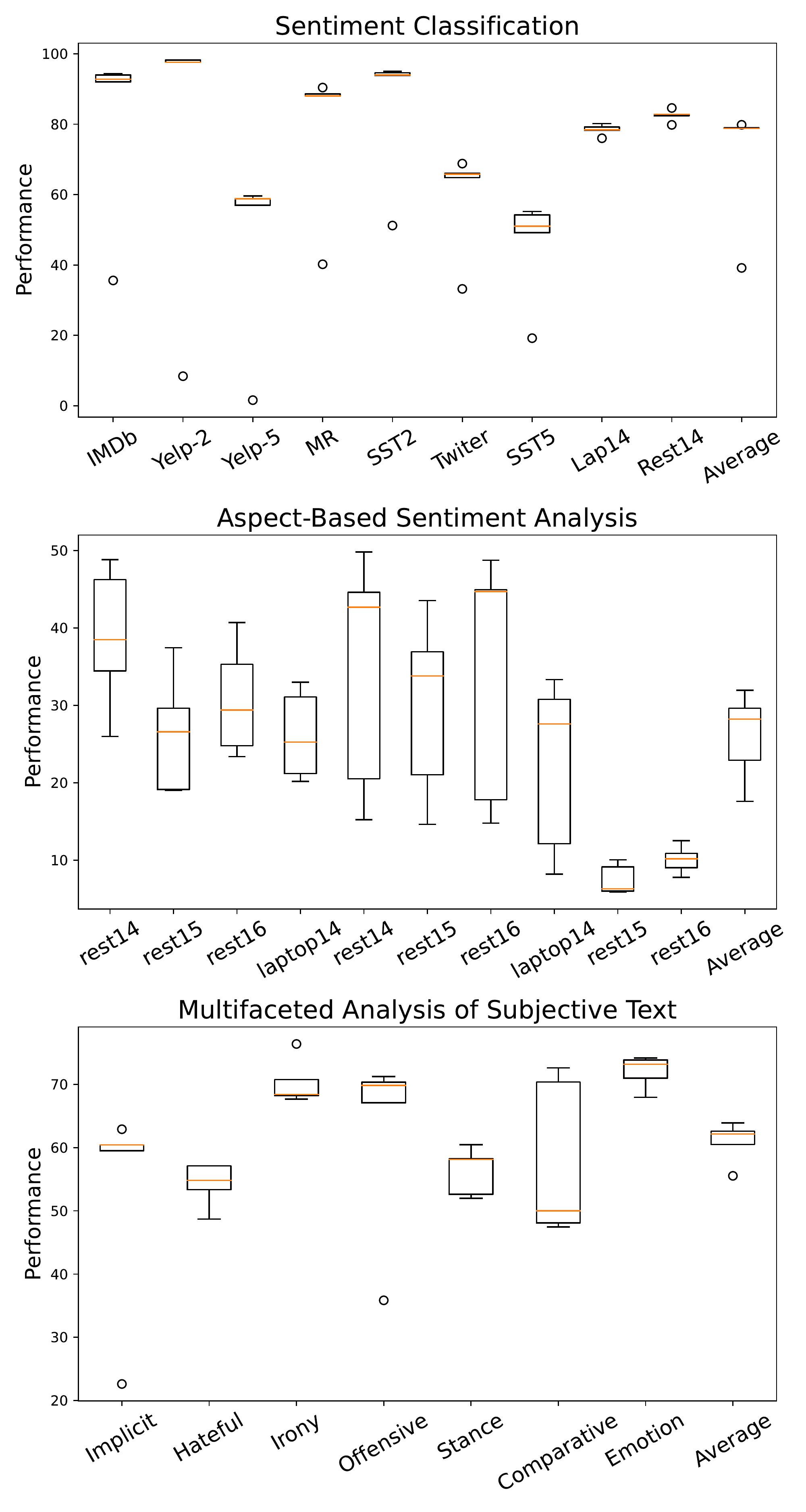}
    \caption{Sensitivity of different prompt designs on three types of SA tasks. The performance variance of each dataset is given by five different prompts. The circles depicted in the figure represent outlier data points.}
    \label{fig:boxplot}
\end{figure}

The design of suitable prompts is critical when leveraging large language models for specific tasks. The different prompt designs have been shown to even lead to large performance variance \cite{nips21-prompt-varaince, acl22-prompt-varaince}. To investigate the impact of such sensitivity on SA tasks, we further construct an additional five prompts for each task, then conduct experiments with \texttt{ChatGPT} to evaluate the variations in performance. 

\begin{figure}[h]
    \centering
    \includegraphics[width=\linewidth]{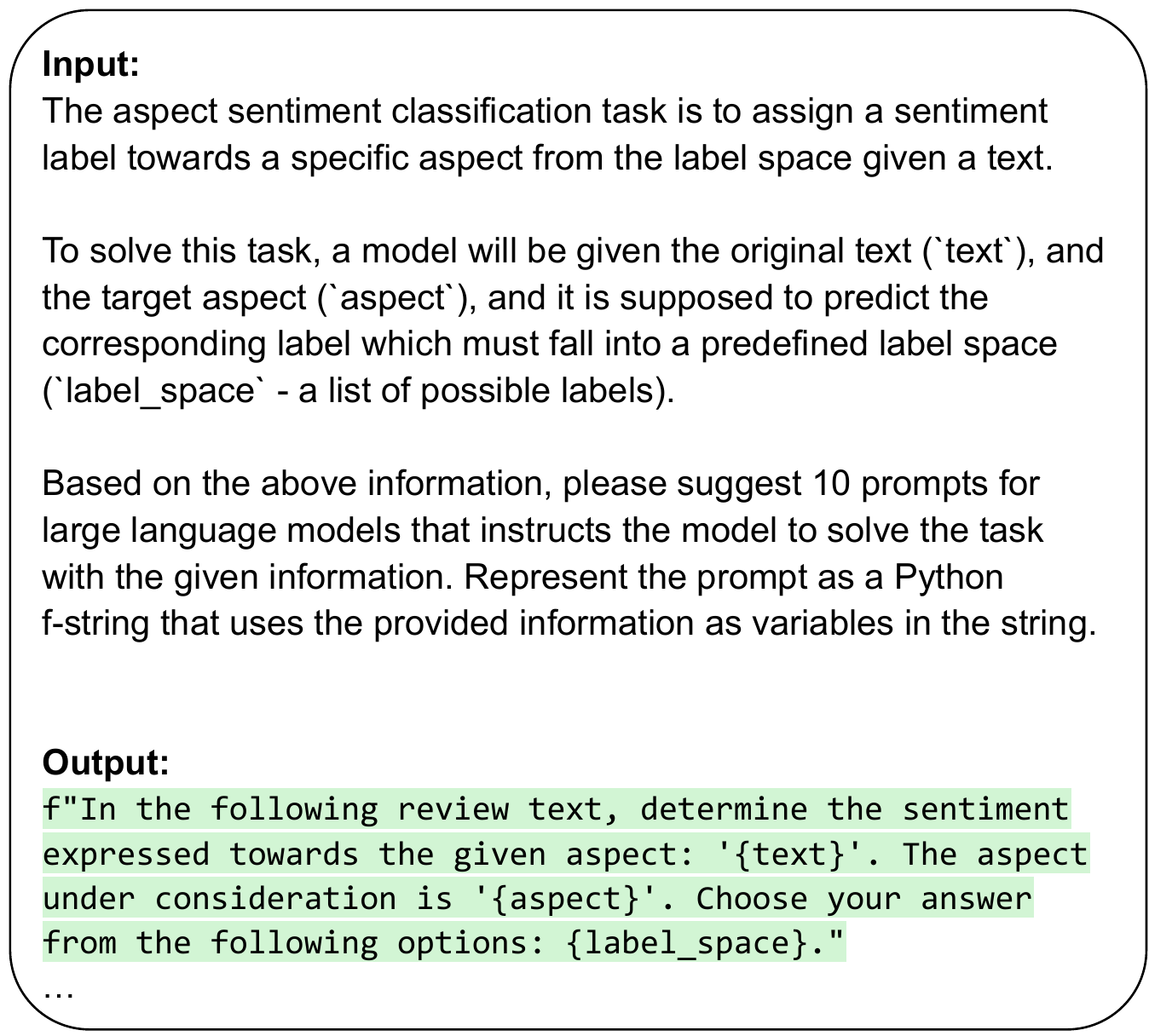}
    \caption{Example prompts generated by GPT-4 for the aspect-level SC task. The first generated prompt is shown for illustrative purposes, and subsequent prompts are not included for brevity.}
    \label{fig:p4p}
\end{figure}

We take GPT-4 \cite{gpt4-report} for such prompt generation\footnote{We also conduct preliminary experiments with \texttt{ChatGPT}, however, it struggles to understand such complicated instructions, thus failing to produce satisfactory prompts.}, which has shown to be effective to generate prompts or instruction-following data \cite{gpt4-instruction-tuning}. This can also alleviate the potential bias of manually written prompts. Specifically, we provide the task description, format requirement (similar to those described in Sec \ref{sec:prompt-strategy}), and an instruction to require it to generate several prompts, representing as Python f-strings. We also optionally provide some input-target pairs to help the model better grasp the goals of the task. We present an example prompt in Figure~\ref{fig:p4p}, using the aspect-level SC task for illustration.

The results of \texttt{ChatGPT} with the five different prompts are depicted in Figure~\ref{fig:boxplot}, in the format of the boxplot. It can be noticed that the impact of different prompts on performance varies from task to task. For SC tasks, the choice of prompt appears to have less effect, e.g., the boxes in the top figure are usually quite concentrated. However, for tasks necessitating structured, fine-grained output, the performance can vary significantly depending on the design of the prompt, as illustrated in the middle figure for ABSA tasks. Interestingly, despite the simplicity of SC tasks, the model still demonstrates sensitivity to certain prompts, with noticeable outliers for some SC datasets (i.e., circles in the figure).

\subsection{Human Evaluation}
\begin{table}[!t]
    \centering
    \resizebox{0.9\linewidth}{!}{
    \begin{tabular}{lccc}
    \toprule
    Task & Comparison & Strict & Relaxed \\
    \midrule
    UABSA & 43.33 & 58.33 & 68.33 \\
    ASTE & 26.67 & 58.33 & 63.33  \\
    ASQP & 10.00 & 26.67 & 40.00 \\
    \bottomrule
    \end{tabular}}
    \caption{Human evaluation under three scenarios: 1) \textbf{comparison} between label and predictions; 2) \textbf{strict} judgment based on annotation rules; 3) \textbf{relaxed} judgment based on prompt only. Note that evaluations are assessed at sentence level, the results are not directly comparable to results in Table~\ref{tab:zero-shot} and Tabel~\ref{tab:few-shot}.}
    \label{tab:human-eval}
\end{table}

The results in Table~\ref{tab:zero-shot} are computed based on automatic evaluation metrics with model predictions. Nevertheless, the generative nature of LLMs can sometimes result in invalid predictions, where the output does not adhere to the required format. This issue is particularly noticeable for ABSA tasks that require structured output from the model. While LLMs seem to underperform, for instance, producing only half the performance of the fine-tuned T5 model on ABSA tasks in Table~\ref{tab:zero-shot}, this poses a natural question: does this performance gap truly reflect the inferiority of LLMs?

We conduct a human evaluation to further investigate such results. We employ three scenarios: 1) comparison: an annotator is asked for comparing a label and prediction pair without prior knowledge of their identities and subsequently required to determine which is superior, or if they are equivalent. We then compute the ratio of acceptance rate with the number of samples where the prediction is equivalent or better than the label; 2) strict: an annotator is first instructed to fully understand the original annotation rules, and then judge whether the prediction is correct or not; 3) relaxed: an annotator (without much prior knowledge in ABSA) is directly asked to judge the goodness of the prediction, only given the same prompt as the LLMs take during the inference. We sample 15 examples from each dataset and provide a total of 150 predictions to three annotators. 

The acceptance ratios under three scenarios are presented in Table~\ref{tab:human-eval}. Upon human evaluation, we observe that the models generally perform better compared to automated evaluations. This suggests that the models are capable of tackling the task but may fail to conform to the required format. With more relaxed requirements, such as when a human is only presented with the prompt as the LLMs, the acceptance ratio increases. However, even under the ``relaxed'' evaluation conditions, the performance is still not satisfactory, indicating that LLMs still struggle to tackle such fine-grained sentiment information.

\subsection{Few-shot Results}

\begin{table*}[!t]
    \centering
    \resizebox{\linewidth}{!}{
    \begin{tabular}{cl|ccc|ccc|cc}
    \toprule
    \multirow{2}{*}{Task} & \multirow{2}{*}{Dataset} & \multicolumn{3}{c}{1-shot} & \multicolumn{3}{c}{5-shot} & \multicolumn{2}{c}{10-shot}  \\
    \cmidrule(lr){3-5} \cmidrule(lr){6-8} \cmidrule(lr){9-10} 
    & & \texttt{Flan-UL2} & \texttt{ChatGPT} & \texttt{T5}$_{large}$ & \texttt{Flan-UL2} &\texttt{ChatGPT} & \texttt{T5}$_{large}$ & \texttt{ChatGPT} &  \texttt{T5}$_{large}$ \\
    \midrule
    \multicolumn{10}{c}{\textit{Sentiment Classification (SC)}} \\
    \midrule
    \rowcolor{lightgray}
     & IMDb & NA & $95.33_{0.50}$ & $77.20_{10.74}$ & NA & NA & $90.00_{2.03}$ & NA &  $91.80_{1.44}$ \\
    \rowcolor{lightgray}
    & Yelp2 & NA & $97.60_{0.92}$ & $86.60_{5.56}$ & NA & NA & $92.40_{0.00}$ & NA &  $90.87_{1.63}$ \\
    \rowcolor{lightgray}
    \multirow{-3}{*}{\parbox{1.2cm}{Document-Level}}& Yelp5 & NA & $51.47_{2.50}$ & $36.47_{4.40}$ & NA & NA & $44.53_{3.19}$ & NA &  $50.60_{0.53}$ \\
    \multirow{4}{*}{\parbox{1.2cm}{Sentence-Level}} & MR & $92.87_{0.23}$ & $91.60_{0.40}$ & $72.87_{9.15}$ & $93.80_{0.00}$ & $90.20_{0.53}$ & $85.67_{1.62}$ & $87.53_{3.44}$ &  $86.60_{1.22}$ \\
    & SST2 & $97.00_{0.20}$ & $94.87_{0.81}$ & $59.33_{2.89}$ & $97.40_{0.20}$ & $95.27_{0.46}$ & $91.40_{3.36}$ & $90.93_{3.72}$ &  $94.60_{0.72}$ \\
    & Twitter & $47.53_{0.31}$ & $66.47_{1.62}$ & $28.33_{7.96}$ & $47.93_{0.31}$ & $64.33_{1.40}$ & $53.20_{4.65}$ & $62.73_{0.81}$ &  $56.60_{3.14}$ \\
    & SST5 & $51.80_{0.92}$ & $51.87_{0.76}$ & $26.67_{1.10}$ & NA & $51.00_{3.27}$ & $39.00_{1.25}$ & $47.60_{1.25}$ &  $40.27_{4.84}$ \\
    \rowcolor{lightgray}
     & Lap14 &  $77.80_{0.35}$ & $78.60_{3.14}$ & $65.47_{1.10}$ & $78.13_{0.42}$ & $76.27_{2.37}$ & $69.13_{1.50}$ & $76.67_{2.41}$ &  $74.40_{0.87}$ \\
    \rowcolor{lightgray}
    \multirow{-2}{*}{\parbox{1.2cm}{Aspect-Level}}& Rest14 & $84.87_{1.03}$ & $84.53_{0.64}$ & $52.47_{19.00}$ & $86.20_{0.92}$ & $74.87_{7.40}$ & $75.80_{0.20}$ & $74.20_{4.13}$ &  $70.47_{1.70}$ \\
    \midrule
    \multicolumn{10}{c}{\textit{Aspect-based Sentiment Analysis (ABSA)}} \\
    \midrule
    \rowcolor{lightgray}
    & Rest14 & $16.67_{2.90}$ & $63.62_{0.89}$ & $18.43_{4.17}$ & NA & $62.40_{1.02}$ & $36.55_{1.92}$ & $63.30_{1.21}$ &  $44.07_{2.19}$ \\
    \rowcolor{lightgray}
    & Rest15 & $16.50_{1.81}$ & $49.35_{2.53}$ & $18.04_{3.89}$ & NA & $52.18_{1.56}$ & $29.95_{0.35}$ & $52.85_{0.75}$ &  $38.96_{1.44}$ \\
    \rowcolor{lightgray}
    & Rest16 & $17.98_{2.10}$ & $56.50_{2.34}$ &$15.86_{4.38}$ & NA & $57.74_{0.39}$ & $32.32_{3.43}$ & $59.22_{2.00}$ &  $46.62_{4.28}$ \\
    \rowcolor{lightgray}
    \multirow{-4}{*}{\parbox{1.2cm}{UABSA}} & Laptop14 & $13.29_{0.88}$ & $40.82_{4.61}$ & $10.47_{2.30}$ & NA & $42.67_{0.12}$ & $20.00_{2.22}$ & $44.70_{1.36}$ &  $28.38_{0.89}$ \\
    \multirow{4}{*}{\parbox{1.2cm}{ASTE}} & Rest14 & $9.26_{1.75}$ & $44.92_{3.53}$ & $5.62_{4.35}$ & NA & $50.75_{5.93}$ & $25.00_{4.09}$ & $54.11_{2.98}$ &  $33.17_{1.21}$ \\
     & Rest15 & $9.31_{0.43}$ & $47.30_{1.96}$ & $9.19_{1.15}$ & NA & $49.99_{4.34}$ & $27.44_{1.26}$ & $48.11_{0.78}$ &  $32.28_{2.29}$ \\
     & Rest16  & $11.81_{1.99}$ & $50.09_{4.28}$ & $9.48_{8.84}$ & NA & $51.30_{0.47}$ & $26.44_{2.52}$ & $53.60_{4.51}$ &  $32.14_{4.38}$ \\
     & Laptop14 & $5.19_{1.54}$ & $35.49_{3.38}$ & $2.94_{2.14}$ & NA & $42.56_{1.78}$ & $15.52_{3.14}$ & $44.74_{2.36}$ & $21.95_{3.50}$ \\
    \rowcolor{lightgray}
      & Rest15 & NA & $30.15_{1.48}$ & $8.69_{0.95}$ & NA & $31.21_{1.94}$ & $13.75_{0.78}$ & $30.92_{2.78}$ &  $14.87_{1.06}$ \\
    \rowcolor{lightgray}
    \multirow{-2}{*}{\parbox{1.2cm}{ASQP}}  & Rest16  & NA & $31.98_{2.06}$ & $2.53_{2.14}$ & NA & $38.01_{2.28}$ & $14.40_{4.76}$ & $40.15_{1.49}$  & $19.23_{1.42}$ \\
    \midrule
    \multicolumn{10}{c}{\textit{Multifaceted Analysis of Subjective Text (MAST)}} \\
    \midrule
    \rowcolor{lightgray}
    Implicit & Lap+Res & $49.40_{0.79}$ & $65.08_{4.89}$ & $34.01_{10.13}$ & $50.91_{1.17}$ & $59.58_{5.01}$ & $46.53_{4.12}$ & $59.73_{1.85}$ &$52.56_{9.98}$\\
    Hate & HatEval & $64.76_{0.97}$ & $55.88_{8.17}$ & $25.77_{3.17}$ & $64.12_{3.32}$ & $50.46_{1.57}$ & $49.89_{5.29}$ & $57.96_{3.34}$ &  $52.54_{3.03}$ \\
    \rowcolor{lightgray}
    Irony & Irony18 & $81.78_{0.87}$ & $79.57_{2.76}$ & $38.23_{10.72}$ & $82.32_{0.45}$ & $84.28_{1.30}$ & $57.69_{7.55}$ & $80.16_{1.47}$ &  $58.90_{2.40}$ \\
    Offensive & OffensEval & $77.29_{0.47}$ & $72.75_{1.63}$ & $17.67_{7.35}$ & $78.01_{1.14}$ & $72.54_{1.34}$ & $49.19_{1.26}$ & $70.21_{3.33}$ &  $49.97_{5.66}$ \\
    \rowcolor{lightgray}
    Stance & Stance16 & $67.75_{1.96}$ & $59.31_{1.81}$ & $33.37_{4.22}$ & $70.49_{0.80}$ & $53.53_{5.04}$ & $35.15_{3.78}$ & $43.15_{5.33}$ &  $36.94_{1.75}$ \\ 
    Comparative & CS19 & $86.62_{1.10}$ & $73.99_{2.96}$ & $46.39_{11.98}$ & $87.26_{1.10}$ & $68.79_{3.32}$ & $70.28_{4.03}$ & $68.26_{3.83}$ &  $71.87_{2.07}$ \\
    \rowcolor{lightgray}
    Emotion & Emotion20 & $71.05_{0.73}$ & $72.59_{2.01}$ & $43.16_{9.98}$ & $69.85_{2.02}$ & $74.30_{2.41}$ & $65.08_{4.23}$ & $69.88_{1.34}$ &  $71.60_{0.55}$ \\
     
    \bottomrule
    \end{tabular}}
    \caption{Few-shot performance of various sentiment analysis tasks. All the results are reported with average and standard deviation in 3 runs. "NA" denotes infeasible experiments due to limited sequence length.}
    \label{tab:few-shot}
\end{table*}

\begin{figure*}[t]
    \centering
    \includegraphics[width=\linewidth]{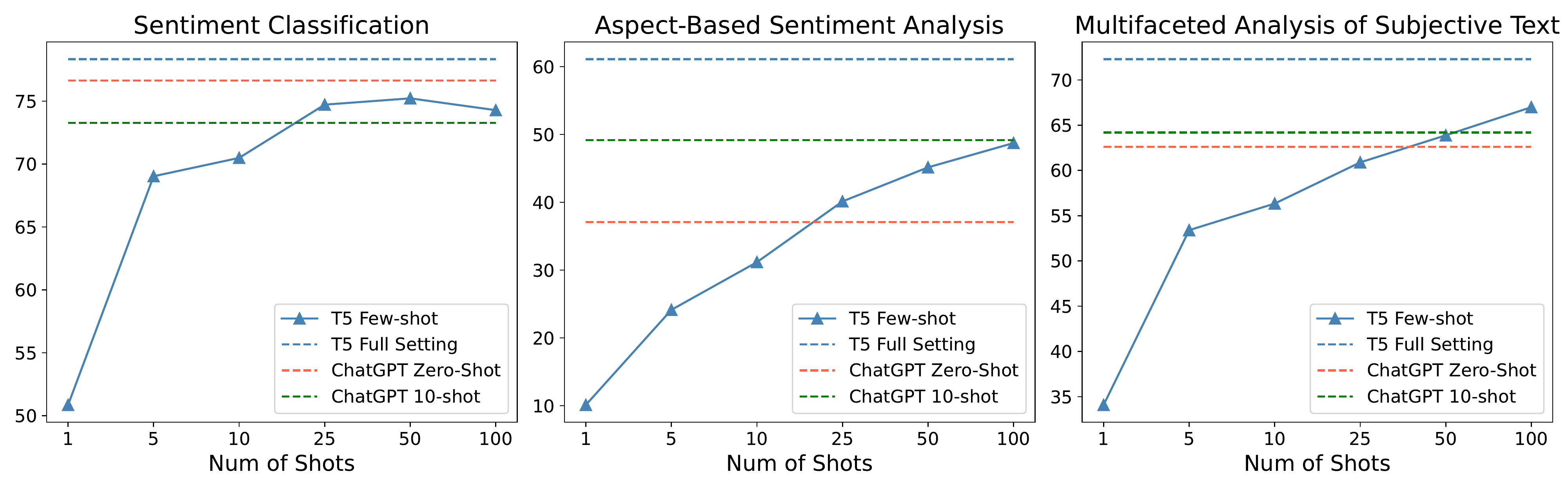}
    \caption{Averaged few-shot results on all datasets for each task type with an increasing number of different shots. Results of ChatGPT zero-shot and T5 full setting are also shown for easy comparison. }
    \label{fig:few-shot}
\end{figure*}

We also conduct few-shot experiments to assess whether LLMs or SLMs perform better when only a limited number of examples for a sentiment analysis task are available. The results of these experiments are summarized in Table~\ref{tab:few-shot}. We consider three K-shot settings: 1-shot, 5-shot, and 10-shot. For each setting, we sample K examples for each sentiment type (with the exception of the ASQP task, where we sample K examples for each aspect category). These sampled examples serve as in-context learning samples for LLMs and training data for SLMs. We have the following findings:

\paragraph{LLMs surpass SLMs under varied few-shot settings}
Across all three few-shot settings, LLMs, whether it is \texttt{ChatGPT} or \texttt{Flan-UL2}, consistently outperform smaller language models \texttt{T5} in almost all cases. This advantage becomes more obvious for ABSA tasks, which require the model to output structured sentiment information. SLMs significantly lag behind LLMs under such requirements, possibly due to the difficulty of learning such patterns with limited data. To delve deeper into their respective strengths and limitations, we gradually increase the value of K in the few-shot settings\footnote{We only report results for SLMs here, as LLMs frequently encounter a context length limit, making them unsuitable for larger K values without specific handling.}, and present the results for \texttt{T5} in Figure~\ref{fig:few-shot}. It becomes apparent that even with a 10-shot setting, \texttt{ChatGPT} sets a robust baseline that requires \texttt{T5} to utilize nearly five to ten times more data to achieve comparable performance.

\paragraph{SLMs show consistent improvements across most tasks with more shots}
As the number of shots increases, SLMs consistently exhibit substantial improvements in various SA tasks. This is in line with our expectations and shows the ability of SLMs to effectively leverage a greater number of examples, thereby achieving better performance.
The task complexity can also be observed from Figure~\ref{fig:few-shot}, where the performance of the \texttt{T5} model begins to gradually plateau for sentiment classification tasks. However, for ABSA and MAST tasks, the performance continues to grow sharply, indicating that these tasks require comparatively more data to capture their underlying patterns.

\paragraph{Increasing shots for LLMs brings different impacts on different tasks}
The impact of increasing shots on LLMs' performance varies from task to task. For relatively easier tasks like SC, the incremental benefit of additional shots for LLMs is less obvious. Moreover, some datasets such as MR and Twitter, along with stance and comparative tasks, even show hindered performance with an increase in the number of shots. This may be due to the consequence of dealing with overly long contexts that could mislead the LLMs.
However, for ABSA tasks, which demand a deeper understanding and precise output format, increasing the number of shots greatly boosts LLM performance. This suggests that the utility of extra examples is not a silver bullet for all tasks but varies depending on the complexity of the task.

\section{\textsc{SentiEval} Benchmark}
\subsection{Rethinking SA Capability Evaluation}
We have conducted extensive experiments to evaluate LLMs' SA capability in the above sections, where we notice some common flaws regarding the current evaluation practice along the way. 

\paragraph{Call for more comprehensive evaluation}
Most of the current evaluations tend to focus narrowly on specific SA tasks or datasets \cite{llm-sa-chatgpt, llm-sa-xiarui}. While these assessments can provide useful insights into certain aspects of an LLM's sentiment analysis competence, they inherently fall short of capturing the full breadth and depth of the model's capabilities. Such limitation not only reduces the overall reliability of the assessment results but also limits the scope of understanding the model's adaptability to diverse SA scenarios. For example, a model with satisfactory sentiment classification ability does not guarantee its performance in detecting hateful speech. Therefore, we attempt to provide a holistic evaluation across a wide range of SA tasks in this work and call for a more comprehensive evaluation on a wider range of SA tasks in the future.

\paragraph{Appeal for more natural ways to interact with the models} 
Conventional sentiment analysis tasks are often structured as a single sentence paired with its corresponding sentiment label. This format, while facilitating the learning of the mapping relationship between the text and its sentiment label, may not optimally suit LLMs, which are typically text generation models. In practice, users exhibit varied writing styles, leading to diverse ways of communicating their requirements to LLMs to solve their SA tasks. It is thus critical to account for these diverse expressions in the evaluation process to reflect more realistic use cases. This ensures the evaluation results mirror real-world interactions, offering more reliable and applicable insights.

\paragraph{Sensitivity on Prompt Design} 
As shown in Sec~\ref{sec:sensitity-prompt}, variations in prompt design can substantially influence the performance of \texttt{ChatGPT}, even on some seemingly simple sentiment classification tasks. Such nuanced sensitivity associated with prompt design introduces challenges when attempting to fairly and stably test the SA capabilities of LLMs. This challenge is further amplified when various studies employ distinct prompts for different SA tasks across a range of LLMs. The inherent bias associated with prompt design complicates the fair comparison of different models using the same prompt, as a single prompt may not be universally appropriate to reflect all models' capabilities.

\subsection{\textsc{SentiEval}: Construction}
To mitigate the limitations when assessing LLMs' SA capability discussed above, we propose the \textsc{SentiEval} benchmark for better \textbf{senti}ment analysis \textbf{eval}uation in the era of large language models. 

The main idea of \textsc{SentiEval} is to: 
1) break the boundary between individual sentiment analysis tasks to establish a unified testing benchmark, providing a more comprehensive assessment of a model's sentiment analysis proficiency, rather than emphasizing on specific aspects;
2) test the model using natural language instructions presented in various styles. This mimics the real use case when humans interact with the model with natural languages for solving SA tasks, instead of purely learning text-label mapping;
3) equip the benchmark with diverse but fixed instructions, making performance comparisons more stable and reliable across different LLMs and studies. By setting a consistent benchmark, it allows for an equitable comparison that is less subject to prompt variation.

Specifically, besides the five prompts generated by GPT-4 in Sec~\ref{sec:sensitity-prompt}, we further manually write five additional prompts for each task. Therefore, each task will have ten candidate prompts in total. Then for each data sample of all tasks, we randomly select one prompt and combine it with the text to form a complete query for the model. Additionally, we also randomly decide (with a 50\% percent chance) whether to put some few-shot examples with the current prompt. In the end, each data sample contains the original text, the instruction for a specific task, and optional few-shot examples.

\subsection{\textsc{SentiEval}: Re-evaluate}
\begin{table}[!t]
    \centering
    \resizebox{\linewidth}{!}{
    \begin{tabular}{lcccc}
    \toprule
     & \texttt{Flan-T5} & \texttt{Flan-UL2} & \texttt{text-003} & \texttt{ChatGPT} \\
    \midrule
    \textsc{SentiEval} & 29.07 & 38.82 & 36.64 & 47.55\\
    \midrule
    SC  & 54.22 & 63.13 & 60.11 & 72.73 \\
    ABSA  & 0.00 & 0.09 & 11.66 & 14.77 \\
    MAST & 34.21 & 58.35 & 38.48 & 57.71 \\
    \bottomrule
    \end{tabular}}
    \caption{Results on the \textsc{SentiEval} benchmark of different LLMs. Predictions are evaluated with the exact match of the label.}
    \label{tab:bigsent}
\end{table}

After constructing the \textsc{SentiEval} benchmark, we revisit the evaluation of the various LLMs outlined in Sec~\ref{sec:llms} against this benchmark. We report the results in Table~\ref{tab:bigsent}, which are the exact match scores between the labels and predictions. Although the new benchmark does not treat each task separately, we further report the results of different tasks for investigations. 

From Table~\ref{tab:bigsent}, we can see the performance gap between different models remains similar to previous zero-shot and few-shot experimental results. To achieve a good performance, it necessitates the model's understanding of varying styles of instructions (i.e., different prompt designs). It also demands the model's compliance with the required format, or adaptation to the pattern set by few-shot examples, thus posing greater challenges. We can see \texttt{ChatGPT} sets a strong performance baseline, distinguishing itself from other LLMs, and showing its strong SA capability and instruction-following ability.  Overall, there is still much room for the LLMs to improve on this benchmark in the future, especially for more complicated tasks such as ABSA and MAST tasks.

\section{Discussions}
\subsection{LLMs for SA in Practice}
In this study, we carry out a comprehensive evaluation of various large language models across a range of sentiment analysis tasks. The experimental results lead us to several primary findings and recommendations for practical SA application:
\begin{itemize}
    \item For simple SA tasks such as binary or trinary sentiment classification, LLMs can already serve as effective solutions. Even in a zero-shot setting, their performance can match or surpass fine-tuned smaller language models, and with little sensitivity to different prompt designs (as shown in Sec \ref{sec:sensitity-prompt}).
    \item When annotation resources are scarce, LLMs remain a good choice due to their superior few-shot in-context learning performance compared to SLMs trained on the same limited data. However, the restricted context length of LLMs can limit their use case, particularly in document-level tasks where SLMs might be more suitable.
    \item For tasks requiring structured sentiment output, like aspect-based sentiment analysis tasks, LLMs might not be the best option. They tend to lag behind SLMs in both automatic and human evaluations, and performance can vary significantly with different prompt designs.
    \item Larger models do not always guarantee superior performance, for instance, \texttt{Flan-UL2} often performs comparably to the GPT-3.5 series of models, despite being much smaller in size. This suggests that employing instruction-tuning to attain a reasonably sized model may suffice for practical SA applications.
\end{itemize}

\subsection{SA Challenges for LLMs}
With the advancement of LLMs, many SA tasks can be claimed to be solved such as binary sentiment classification, as we saw from the experimental results. However, does it mean sentiment analysis has reached its maturity in the era of LLMs? We discuss some remaining challenges that we think still pose great difficulties.

\paragraph{Understanding Complex Linguistic Nuances and Cultural Specificity} Sentiment is often shaded with nuance and subtlety. Developing models capable of understanding such subtleties in language, such as sarcasm, irony, humor, and specific cultural idioms or expressions is still challenging. They often depend on the context and shared cultural background knowledge or even specific human experiences. For example, on Chinese social media, a comment ``\cchar{您说的都对}'' (English translation: ``You are right about everything you said'' with ``You'' in a respectful tone) may not necessarily indicate agreement but can be used ironically. However, this linguistic phenomenon may require familiarity with social media to interpret correctly.

\paragraph{Extracting fine-grained and structured sentiment information} As can be seen from the results, requiring the models to generate structured fine-grained information, i.e., the ABSA tasks, is still challenging for the models. However, such information can be useful to quickly summarize large-scale information to produce a more organized digest, especially since the long context is still a limitation for many LLMs. Also, distinguishing more precise emotional states or intensities of sentiment for more detailed analysis is also challenging but worth exploring. 

\paragraph{Real-Time Adaptation for Evolving Sentiment Analysis}
Sentiments and expressions constantly evolve, particularly on platforms like social media. This leads to the continual emergence of new idioms and sentiment-caring expressions. It thus demands the sentiment analysis models to adapt and learn from these evolving trends to accurately interpret the embedded sentiments. However, one of the major limitations of current LLMs lies in their lack of flexibility in fine-tuning or re-training. This issue restricts their capability to keep up with the fast-paced evolution of language and sentiment, resulting in outdated or inaccurate sentiment analysis. Therefore, a critical research direction involves developing methods for rapid and effective model updates to ensure real-time and accurate sentiment analysis.

\section{Conclusions}
In this study, we conduct a systematic evaluation of various sentiment analysis tasks using LLMs, which helps better understand their capabilities in sentiment analysis problems. Experimental results reveal that while LLMs perform quite well on simpler tasks in a zero-shot setting, they struggle with more complex tasks. In a few-shot learning context, LLMs consistently outperform SLMs, suggesting their potential in scenarios where annotation resources are scarce. This work also highlights the limitations of current evaluation practices and then introduces the \textsc{SentiEval} benchmark as a more comprehensive and realistic evaluation tool.

Overall, large language models have opened new avenues for sentiment analysis. While some traditional SA tasks have achieved near-human performance, a comprehensive understanding of human sentiment, opinion, and other subjective feelings remains a long way to pursue. The powerful text comprehension capabilities of LLMs offer effective tools and exciting research directions for the exploration of sentiment analysis in the LLM era.

\bibliography{custom}

\begin{thebibliography}{57}
\expandafter\ifx\csname natexlab\endcsname\relax\def\natexlab#1{#1}\fi

\bibitem[{Bang et~al.(2023)Bang, Cahyawijaya, Lee, Dai, Su, Wilie, Lovenia, Ji,
  Yu, Chung, Do, Xu, and Fung}]{chatgpt-evaluate}
Yejin Bang, Samuel Cahyawijaya, Nayeon Lee, Wenliang Dai, Dan Su, Bryan Wilie,
  Holy Lovenia, Ziwei Ji, Tiezheng Yu, Willy Chung, Quyet~V. Do, Yan Xu, and
  Pascale Fung. 2023.
\newblock \href {https://doi.org/10.48550/arXiv.2302.04023} {A multitask,
  multilingual, multimodal evaluation of chatgpt on reasoning, hallucination,
  and interactivity}.
\newblock \emph{CoRR}, abs/2302.04023.

\bibitem[{Barbieri et~al.(2020)Barbieri, Camacho{-}Collados, Anke, and
  Neves}]{tweeteval}
Francesco Barbieri, Jos{\'{e}} Camacho{-}Collados, Luis~Espinosa Anke, and
  Leonardo Neves. 2020.
\newblock \href {https://doi.org/10.18653/v1/2020.findings-emnlp.148}
  {Tweeteval: Unified benchmark and comparative evaluation for tweet
  classification}.
\newblock In \emph{Findings of the Association for Computational Linguistics:
  {EMNLP} 2020, Online Event, 16-20 November 2020}, pages 1644--1650.

\bibitem[{Basile et~al.(2019)Basile, Bosco, Fersini, Nozza, Patti, Pardo,
  Rosso, and Sanguinetti}]{hate}
Valerio Basile, Cristina Bosco, Elisabetta Fersini, Debora Nozza, Viviana
  Patti, Francisco Manuel~Rangel Pardo, Paolo Rosso, and Manuela Sanguinetti.
  2019.
\newblock \href {https://doi.org/10.18653/v1/s19-2007} {Semeval-2019 task 5:
  Multilingual detection of hate speech against immigrants and women in
  twitter}.
\newblock In \emph{Proceedings of the 13th International Workshop on Semantic
  Evaluation, SemEval@NAACL-HLT 2019, Minneapolis, MN, USA, June 6-7, 2019},
  pages 54--63. Association for Computational Linguistics.

\bibitem[{Brown et~al.(2020)Brown, Mann, Ryder, Subbiah, Kaplan, Dhariwal,
  Neelakantan, Shyam, Sastry, Askell, Agarwal, Herbert{-}Voss, Krueger,
  Henighan, Child, Ramesh, Ziegler, Wu, Winter, Hesse, Chen, Sigler, Litwin,
  Gray, Chess, Clark, Berner, McCandlish, Radford, Sutskever, and
  Amodei}]{gpt3-paper}
Tom~B. Brown, Benjamin Mann, Nick Ryder, Melanie Subbiah, Jared Kaplan,
  Prafulla Dhariwal, Arvind Neelakantan, Pranav Shyam, Girish Sastry, Amanda
  Askell, Sandhini Agarwal, Ariel Herbert{-}Voss, Gretchen Krueger, Tom
  Henighan, Rewon Child, Aditya Ramesh, Daniel~M. Ziegler, Jeffrey Wu, Clemens
  Winter, Christopher Hesse, Mark Chen, Eric Sigler, Mateusz Litwin, Scott
  Gray, Benjamin Chess, Jack Clark, Christopher Berner, Sam McCandlish, Alec
  Radford, Ilya Sutskever, and Dario Amodei. 2020.
\newblock \href
  {https://proceedings.neurips.cc/paper/2020/hash/1457c0d6bfcb4967418bfb8ac142f64a-Abstract.html}
  {Language models are few-shot learners}.
\newblock In \emph{Advances in Neural Information Processing Systems 33: Annual
  Conference on Neural Information Processing Systems 2020, NeurIPS 2020}.

\bibitem[{Bubeck et~al.(2023)Bubeck, Chandrasekaran, Eldan, Gehrke, Horvitz,
  Kamar, Lee, Lee, Li, Lundberg, Nori, Palangi, Ribeiro, and
  Zhang}]{sparks-agi}
S{\'{e}}bastien Bubeck, Varun Chandrasekaran, Ronen Eldan, Johannes Gehrke,
  Eric Horvitz, Ece Kamar, Peter Lee, Yin~Tat Lee, Yuanzhi Li, Scott~M.
  Lundberg, Harsha Nori, Hamid Palangi, Marco~T{\'{u}}lio Ribeiro, and
  Yi~Zhang. 2023.
\newblock \href {https://doi.org/10.48550/arXiv.2303.12712} {Sparks of
  artificial general intelligence: Early experiments with {GPT-4}}.
\newblock \emph{CoRR}, abs/2303.12712.

\bibitem[{Cai et~al.(2021)Cai, Xia, and Yu}]{acl20-acos}
Hongjie Cai, Rui Xia, and Jianfei Yu. 2021.
\newblock \href {https://doi.org/10.18653/v1/2021.acl-long.29}
  {Aspect-category-opinion-sentiment quadruple extraction with implicit aspects
  and opinions}.
\newblock In \emph{Proceedings of the 59th Annual Meeting of the Association
  for Computational Linguistics and the 11th International Joint Conference on
  Natural Language Processing, {ACL/IJCNLP} 2021}, pages 340--350. Association
  for Computational Linguistics.

\bibitem[{Chowdhery et~al.(2022)Chowdhery, Narang, Devlin, Bosma, Mishra,
  Roberts, Barham, Chung, Sutton, Gehrmann, Schuh, Shi, Tsvyashchenko, Maynez,
  Rao, Barnes, Tay, Shazeer, Prabhakaran, Reif, Du, Hutchinson, Pope, Bradbury,
  Austin, Isard, Gur{-}Ari, Yin, Duke, Levskaya, Ghemawat, Dev, Michalewski,
  Garcia, Misra, Robinson, Fedus, Zhou, Ippolito, Luan, Lim, Zoph, Spiridonov,
  Sepassi, Dohan, Agrawal, Omernick, Dai, Pillai, Pellat, Lewkowycz, Moreira,
  Child, Polozov, Lee, Zhou, Wang, Saeta, Diaz, Firat, Catasta, Wei,
  Meier{-}Hellstern, Eck, Dean, Petrov, and Fiedel}]{palm-paper}
Aakanksha Chowdhery, Sharan Narang, Jacob Devlin, Maarten Bosma, Gaurav Mishra,
  Adam Roberts, Paul Barham, Hyung~Won Chung, Charles Sutton, Sebastian
  Gehrmann, Parker Schuh, Kensen Shi, Sasha Tsvyashchenko, Joshua Maynez,
  Abhishek Rao, Parker Barnes, Yi~Tay, Noam Shazeer, Vinodkumar Prabhakaran,
  Emily Reif, Nan Du, Ben Hutchinson, Reiner Pope, James Bradbury, Jacob
  Austin, Michael Isard, Guy Gur{-}Ari, Pengcheng Yin, Toju Duke, Anselm
  Levskaya, Sanjay Ghemawat, Sunipa Dev, Henryk Michalewski, Xavier Garcia,
  Vedant Misra, Kevin Robinson, Liam Fedus, Denny Zhou, Daphne Ippolito, David
  Luan, Hyeontaek Lim, Barret Zoph, Alexander Spiridonov, Ryan Sepassi, David
  Dohan, Shivani Agrawal, Mark Omernick, Andrew~M. Dai,
  Thanumalayan~Sankaranarayana Pillai, Marie Pellat, Aitor Lewkowycz, Erica
  Moreira, Rewon Child, Oleksandr Polozov, Katherine Lee, Zongwei Zhou, Xuezhi
  Wang, Brennan Saeta, Mark Diaz, Orhan Firat, Michele Catasta, Jason Wei,
  Kathy Meier{-}Hellstern, Douglas Eck, Jeff Dean, Slav Petrov, and Noah
  Fiedel. 2022.
\newblock \href {https://doi.org/10.48550/arXiv.2204.02311} {Palm: Scaling
  language modeling with pathways}.
\newblock \emph{CoRR}, abs/2204.02311.

\bibitem[{Christiano et~al.(2017)Christiano, Leike, Brown, Martic, Legg, and
  Amodei}]{RLHF}
Paul~F. Christiano, Jan Leike, Tom~B. Brown, Miljan Martic, Shane Legg, and
  Dario Amodei. 2017.
\newblock \href
  {https://proceedings.neurips.cc/paper/2017/hash/d5e2c0adad503c91f91df240d0cd4e49-Abstract.html}
  {Deep reinforcement learning from human preferences}.
\newblock In \emph{Advances in Neural Information Processing Systems 30: Annual
  Conference on Neural Information Processing Systems 2017, December 4-9, 2017,
  Long Beach, CA, {USA}}.

\bibitem[{Chung et~al.(2022)Chung, Hou, Longpre, Zoph, Tay, Fedus, Li, Wang,
  Dehghani, Brahma, Webson, Gu, Dai, Suzgun, Chen, Chowdhery, Narang, Mishra,
  Yu, Zhao, Huang, Dai, Yu, Petrov, Chi, Dean, Devlin, Roberts, Zhou, Le, and
  Wei}]{flan-t5-paper}
Hyung~Won Chung, Le~Hou, Shayne Longpre, Barret Zoph, Yi~Tay, William Fedus,
  Eric Li, Xuezhi Wang, Mostafa Dehghani, Siddhartha Brahma, Albert Webson,
  Shixiang~Shane Gu, Zhuyun Dai, Mirac Suzgun, Xinyun Chen, Aakanksha
  Chowdhery, Sharan Narang, Gaurav Mishra, Adams Yu, Vincent~Y. Zhao, Yanping
  Huang, Andrew~M. Dai, Hongkun Yu, Slav Petrov, Ed~H. Chi, Jeff Dean, Jacob
  Devlin, Adam Roberts, Denny Zhou, Quoc~V. Le, and Jason Wei. 2022.
\newblock \href {https://doi.org/10.48550/arXiv.2210.11416} {Scaling
  instruction-finetuned language models}.
\newblock \emph{CoRR}, abs/2210.11416.

\bibitem[{Deng et~al.(2023)Deng, Bashlovkina, Han, Baumgartner, and
  Bendersky}]{www23-llm-sa}
Xiang Deng, Vasilisa Bashlovkina, Feng Han, Simon Baumgartner, and Michael
  Bendersky. 2023.
\newblock \href {https://doi.org/10.1145/3543873.3587605} {Llms to the moon?
  reddit market sentiment analysis with large language models}.
\newblock In \emph{Companion Proceedings of the {ACM} Web Conference 2023,
  {WWW} 2023}, pages 1014--1019.

\bibitem[{Hee et~al.(2018)Hee, Lefever, and Hoste}]{irony}
Cynthia~Van Hee, Els Lefever, and V{\'{e}}ronique Hoste. 2018.
\newblock \href {https://doi.org/10.18653/v1/s18-1005} {Semeval-2018 task 3:
  Irony detection in english tweets}.
\newblock In \emph{Proceedings of The 12th International Workshop on Semantic
  Evaluation, SemEval@NAACL-HLT 2018, New Orleans, Louisiana, USA, June 5-6,
  2018}, pages 39--50. Association for Computational Linguistics.

\bibitem[{Hu and Liu(2004)}]{kdd04-sa}
Minqing Hu and Bing Liu. 2004.
\newblock \href {https://doi.org/10.1145/1014052.1014073} {Mining and
  summarizing customer reviews}.
\newblock In \emph{Proceedings of the Tenth {ACM} {SIGKDD} International
  Conference on Knowledge Discovery and Data Mining}, pages 168--177.

\bibitem[{Keung et~al.(2020)Keung, Lu, Szarvas, and
  Smith}]{emnlp20-multilingual-amazon-review}
Phillip Keung, Yichao Lu, Gy{\"{o}}rgy Szarvas, and Noah~A. Smith. 2020.
\newblock \href {https://doi.org/10.18653/v1/2020.emnlp-main.369} {The
  multilingual amazon reviews corpus}.
\newblock In \emph{Proceedings of the 2020 Conference on Empirical Methods in
  Natural Language Processing, {EMNLP} 2020}, pages 4563--4568. Association for
  Computational Linguistics.

\bibitem[{K\"{u}\c{c}\"{u}k and Can(2020)}]{stance-survey}
Dilek K\"{u}\c{c}\"{u}k and Fazli Can. 2020.
\newblock \href {https://doi.org/10.1145/3369026} {Stance detection: A survey}.
\newblock \emph{ACM Comput. Surv.}, 53(1).

\bibitem[{Li et~al.(2021)Li, Zou, Zhang, Zhang, and Wei}]{implicit}
Zhengyan Li, Yicheng Zou, Chong Zhang, Qi~Zhang, and Zhongyu Wei. 2021.
\newblock \href {https://doi.org/10.18653/v1/2021.emnlp-main.22} {Learning
  implicit sentiment in aspect-based sentiment analysis with supervised
  contrastive pre-training}.
\newblock In \emph{Proceedings of the 2021 Conference on Empirical Methods in
  Natural Language Processing, {EMNLP} 2021, Virtual Event / Punta Cana,
  Dominican Republic, 7-11 November, 2021}, pages 246--256.

\bibitem[{Liu(2015)}]{liubing-sa-book}
Bing Liu. 2015.
\newblock \href
  {http://www.cambridge.org/us/academic/subjects/computer-science/knowledge-management-databases-and-data-mining/sentiment-analysis-mining-opinions-sentiments-and-emotions}
  {\emph{Sentiment Analysis - Mining Opinions, Sentiments, and Emotions}}.
\newblock Cambridge University Press.

\bibitem[{Liu et~al.(2021)Liu, Zheng, Demasi, Sabour, Li, Yu, Jiang, and
  Huang}]{acl21-emotional-dialog}
Siyang Liu, Chujie Zheng, Orianna Demasi, Sahand Sabour, Yu~Li, Zhou Yu, Yong
  Jiang, and Minlie Huang. 2021.
\newblock \href {https://doi.org/10.18653/v1/2021.acl-long.269} {Towards
  emotional support dialog systems}.
\newblock In \emph{Proceedings of the 59th Annual Meeting of the Association
  for Computational Linguistics and the 11th International Joint Conference on
  Natural Language Processing, {ACL/IJCNLP} 2021}, pages 3469--3483.

\bibitem[{Lu et~al.(2022)Lu, Bartolo, Moore, Riedel, and
  Stenetorp}]{acl22-prompt-varaince}
Yao Lu, Max Bartolo, Alastair Moore, Sebastian Riedel, and Pontus Stenetorp.
  2022.
\newblock \href {https://doi.org/10.18653/v1/2022.acl-long.556} {Fantastically
  ordered prompts and where to find them: Overcoming few-shot prompt order
  sensitivity}.
\newblock In \emph{Proceedings of the 60th Annual Meeting of the Association
  for Computational Linguistics (Volume 1: Long Papers), {ACL} 2022}, pages
  8086--8098. Association for Computational Linguistics.

\bibitem[{Maas et~al.(2011)Maas, Daly, Pham, Huang, Ng, and Potts}]{imdb}
Andrew~L. Maas, Raymond~E. Daly, Peter~T. Pham, Dan Huang, Andrew~Y. Ng, and
  Christopher Potts. 2011.
\newblock \href {https://aclanthology.org/P11-1015/} {Learning word vectors for
  sentiment analysis}.
\newblock In \emph{The 49th Annual Meeting of the Association for Computational
  Linguistics: Human Language Technologies, Proceedings of the Conference,
  19-24 June, 2011, Portland, Oregon, {USA}}, pages 142--150.

\bibitem[{Mohammad et~al.(2018)Mohammad, Bravo{-}Marquez, Salameh, and
  Kiritchenko}]{emotion-semeval}
Saif~M. Mohammad, Felipe Bravo{-}Marquez, Mohammad Salameh, and Svetlana
  Kiritchenko. 2018.
\newblock \href {https://doi.org/10.18653/v1/s18-1001} {Semeval-2018 task 1:
  Affect in tweets}.
\newblock In \emph{Proceedings of The 12th International Workshop on Semantic
  Evaluation, SemEval@NAACL-HLT 2018, New Orleans, Louisiana, USA, June 5-6,
  2018}, pages 1--17. Association for Computational Linguistics.

\bibitem[{Mohammad et~al.(2016)Mohammad, Kiritchenko, Sobhani, Zhu, and
  Cherry}]{stance}
Saif~M. Mohammad, Svetlana Kiritchenko, Parinaz Sobhani, Xiao{-}Dan Zhu, and
  Colin Cherry. 2016.
\newblock \href {https://doi.org/10.18653/v1/s16-1003} {Semeval-2016 task 6:
  Detecting stance in tweets}.
\newblock In \emph{Proceedings of the 10th International Workshop on Semantic
  Evaluation, SemEval@NAACL-HLT 2016, San Diego, CA, USA, June 16-17, 2016},
  pages 31--41. The Association for Computer Linguistics.

\bibitem[{OpenAI(2023)}]{gpt4-report}
OpenAI. 2023.
\newblock \href {https://doi.org/10.48550/arXiv.2303.08774} {{GPT-4} technical
  report}.
\newblock \emph{CoRR}, abs/2303.08774.

\bibitem[{Ouyang et~al.(2022)Ouyang, Wu, Jiang, Almeida, Wainwright, Mishkin,
  Zhang, Agarwal, Slama, Ray, Schulman, Hilton, Kelton, Miller, Simens, Askell,
  Welinder, Christiano, Leike, and Lowe}]{instruct-gpt}
Long Ouyang, Jeffrey Wu, Xu~Jiang, Diogo Almeida, Carroll~L. Wainwright, Pamela
  Mishkin, Chong Zhang, Sandhini Agarwal, Katarina Slama, Alex Ray, John
  Schulman, Jacob Hilton, Fraser Kelton, Luke Miller, Maddie Simens, Amanda
  Askell, Peter Welinder, Paul~F. Christiano, Jan Leike, and Ryan Lowe. 2022.
\newblock \href
  {http://papers.nips.cc/paper\_files/paper/2022/hash/b1efde53be364a73914f58805a001731-Abstract-Conference.html}
  {Training language models to follow instructions with human feedback}.
\newblock In \emph{NeurIPS}.

\bibitem[{Panchenko et~al.(2019)Panchenko, Bondarenko, Franzek, Hagen, and
  Biemann}]{comparative-CS19}
Alexander Panchenko, Alexander Bondarenko, Mirco Franzek, Matthias Hagen, and
  Chris Biemann. 2019.
\newblock \href {https://doi.org/10.18653/v1/w19-4516} {Categorizing
  comparative sentences}.
\newblock In \emph{Proceedings of the 6th Workshop on Argument Mining,
  ArgMining@ACL 2019, Florence, Italy, August 1, 2019}, pages 136--145.

\bibitem[{Pang and Lee(2005)}]{mr}
Bo~Pang and Lillian Lee. 2005.
\newblock \href {https://aclanthology.org/P05-1015/} {Seeing stars: Exploiting
  class relationships for sentiment categorization with respect to rating
  scales}.
\newblock In \emph{{ACL} 2005, 43rd Annual Meeting of the Association for
  Computational Linguistics, Proceedings of the Conference, 25-30 June 2005,
  University of Michigan, {USA}}, pages 115--124.

\bibitem[{Peng et~al.(2023)Peng, Li, He, Galley, and
  Gao}]{gpt4-instruction-tuning}
Baolin Peng, Chunyuan Li, Pengcheng He, Michel Galley, and Jianfeng Gao. 2023.
\newblock \href {https://doi.org/10.48550/arXiv.2304.03277} {Instruction tuning
  with {GPT-4}}.
\newblock \emph{CoRR}, abs/2304.03277.

\bibitem[{Perez et~al.(2021)Perez, Kiela, and Cho}]{nips21-prompt-varaince}
Ethan Perez, Douwe Kiela, and Kyunghyun Cho. 2021.
\newblock \href
  {https://proceedings.neurips.cc/paper/2021/hash/5c04925674920eb58467fb52ce4ef728-Abstract.html}
  {True few-shot learning with language models}.
\newblock In \emph{Advances in Neural Information Processing Systems 34: Annual
  Conference on Neural Information Processing Systems 2021, NeurIPS 2021},
  pages 11054--11070.

\bibitem[{Pontiki et~al.(2016)Pontiki, Galanis, Papageorgiou, Androutsopoulos,
  Manandhar, AL-Smadi, Al-Ayyoub, Zhao, Qin, De~Clercq, Hoste, Apidianaki,
  Tannier, Loukachevitch, Kotelnikov, Bel, Jim{\'e}nez-Zafra, and
  Eryi{\u{g}}it}]{semeval16}
Maria Pontiki, Dimitris Galanis, Haris Papageorgiou, Ion Androutsopoulos,
  Suresh Manandhar, Mohammad AL-Smadi, Mahmoud Al-Ayyoub, Yanyan Zhao, Bing
  Qin, Orph{\'e}e De~Clercq, V{\'e}ronique Hoste, Marianna Apidianaki, Xavier
  Tannier, Natalia Loukachevitch, Evgeniy Kotelnikov, Nuria Bel,
  Salud~Mar{\'\i}a Jim{\'e}nez-Zafra, and G{\"u}l{\c{s}}en Eryi{\u{g}}it. 2016.
\newblock \href {https://aclanthology.org/S16-1002} {{S}em{E}val-2016 task 5:
  Aspect based sentiment analysis}.
\newblock In \emph{Proceedings of the 10th International Workshop on Semantic
  Evaluation ({S}em{E}val-2016)}, pages 19--30.

\bibitem[{Pontiki et~al.(2015)Pontiki, Galanis, Papageorgiou, Manandhar, and
  Androutsopoulos}]{semeval15}
Maria Pontiki, Dimitris Galanis, Haris Papageorgiou, Suresh Manandhar, and Ion
  Androutsopoulos. 2015.
\newblock \href {https://aclanthology.org/S15-2082} {{S}em{E}val-2015 task 12:
  Aspect based sentiment analysis}.
\newblock In \emph{Proceedings of the 9th International Workshop on Semantic
  Evaluation ({S}em{E}val 2015)}, pages 486--495.

\bibitem[{Pontiki et~al.(2014)Pontiki, Galanis, Pavlopoulos, Papageorgiou,
  Androutsopoulos, and Manandhar}]{semeval14}
Maria Pontiki, Dimitris Galanis, John Pavlopoulos, Harris Papageorgiou, Ion
  Androutsopoulos, and Suresh Manandhar. 2014.
\newblock \href {https://doi.org/10.3115/v1/s14-2004} {Semeval-2014 task 4:
  Aspect based sentiment analysis}.
\newblock In \emph{Proceedings of the 8th International Workshop on Semantic
  Evaluation, SemEval@COLING 2014, Dublin, Ireland, August 23-24, 2014}, pages
  27--35.

\bibitem[{Poria et~al.(2020)Poria, Hazarika, Majumder, and
  Mihalcea}]{tac20-sa-survey}
Soujanya Poria, Devamanyu Hazarika, Navonil Majumder, and Rada Mihalcea. 2020.
\newblock \href {https://doi.org/10.1109/TAFFC.2020.3038167} {Beneath the tip
  of the iceberg: Current challenges and new directions in sentiment analysis
  research}.
\newblock \emph{{IEEE} Trans. Affect. Comput.}

\bibitem[{Pradhan et~al.(2020)Pradhan, Chaturvedi, Tripathi, and
  Sharma}]{offensive-survey}
Rahul Pradhan, Ankur Chaturvedi, Aprna Tripathi, and Dilip~Kumar Sharma. 2020.
\newblock \href
  {https://link.springer.com/chapter/10.1007/978-981-15-0694-9_41} {A review on
  offensive language detection}.
\newblock \emph{Advances in Data and Information Sciences: Proceedings of ICDIS
  2019}, pages 433--439.

\bibitem[{Raffel et~al.(2020)Raffel, Shazeer, Roberts, Lee, Narang, Matena,
  Zhou, Li, and Liu}]{t5-paper}
Colin Raffel, Noam Shazeer, Adam Roberts, Katherine Lee, Sharan Narang, Michael
  Matena, Yanqi Zhou, Wei Li, and Peter~J. Liu. 2020.
\newblock \href {http://jmlr.org/papers/v21/20-074.html} {Exploring the limits
  of transfer learning with a unified text-to-text transformer}.
\newblock \emph{Journal of Machine Learning Research}, 21(140):1--67.

\bibitem[{Rashkin et~al.(2019)Rashkin, Smith, Li, and
  Boureau}]{acl19-empathetic-dialog}
Hannah Rashkin, Eric~Michael Smith, Margaret Li, and Y{-}Lan Boureau. 2019.
\newblock \href {https://doi.org/10.18653/v1/p19-1534} {Towards empathetic
  open-domain conversation models: {A} new benchmark and dataset}.
\newblock In \emph{Proceedings of the 57th Conference of the Association for
  Computational Linguistics, {ACL} 2019}, pages 5370--5381. Association for
  Computational Linguistics.

\bibitem[{Rosenthal et~al.(2017)Rosenthal, Farra, and Nakov}]{twitter-sc}
Sara Rosenthal, Noura Farra, and Preslav Nakov. 2017.
\newblock \href {https://doi.org/10.18653/v1/S17-2088} {Semeval-2017 task 4:
  Sentiment analysis in twitter}.
\newblock In \emph{Proceedings of the 11th international workshop on semantic
  evaluation (SemEval-2017)}, pages 502--518.

\bibitem[{Sailunaz et~al.(2018)Sailunaz, Dhaliwal, Rokne, and
  Alhajj}]{emotion-survey}
Kashfia Sailunaz, Manmeet Dhaliwal, Jon~G. Rokne, and Reda Alhajj. 2018.
\newblock \href {https://doi.org/10.1007/s13278-018-0505-2} {Emotion detection
  from text and speech: a survey}.
\newblock \emph{Soc. Netw. Anal. Min.}, 8(1):28:1--28:26.

\bibitem[{Schmidt and Wiegand(2017)}]{hate-survey}
Anna Schmidt and Michael Wiegand. 2017.
\newblock \href {https://doi.org/10.18653/v1/w17-1101} {A survey on hate speech
  detection using natural language processing}.
\newblock In \emph{Proceedings of the Fifth International Workshop on Natural
  Language Processing for Social Media, SocialNLP@EACL 2017, Valencia, Spain,
  April 3, 2017}, pages 1--10. Association for Computational Linguistics.

\bibitem[{Socher et~al.(2013)Socher, Perelygin, Wu, Chuang, Manning, Ng, and
  Potts}]{sst}
Richard Socher, Alex Perelygin, Jean Wu, Jason Chuang, Christopher~D. Manning,
  Andrew~Y. Ng, and Christopher Potts. 2013.
\newblock \href {https://aclanthology.org/D13-1170/} {Recursive deep models for
  semantic compositionality over a sentiment treebank}.
\newblock In \emph{Proceedings of the 2013 Conference on Empirical Methods in
  Natural Language Processing, {EMNLP} 2013, 18-21 October 2013, Grand Hyatt
  Seattle, Seattle, Washington, USA, {A} meeting of SIGDAT, a Special Interest
  Group of the {ACL}}, pages 1631--1642.

\bibitem[{Tay et~al.(2022)Tay, Dehghani, Tran, Garcia, Bahri, Schuster, Zheng,
  Houlsby, and Metzler}]{ul2-paper}
Yi~Tay, Mostafa Dehghani, Vinh~Q. Tran, Xavier Garcia, Dara Bahri, Tal
  Schuster, Huaixiu~Steven Zheng, Neil Houlsby, and Donald Metzler. 2022.
\newblock \href {https://doi.org/10.48550/arXiv.2205.05131} {Unifying language
  learning paradigms}.
\newblock \emph{CoRR}, abs/2205.05131.

\bibitem[{Touvron et~al.(2023)Touvron, Lavril, Izacard, Martinet, Lachaux,
  Lacroix, Rozi{\`{e}}re, Goyal, Hambro, Azhar, Rodriguez, Joulin, Grave, and
  Lample}]{llama}
Hugo Touvron, Thibaut Lavril, Gautier Izacard, Xavier Martinet, Marie{-}Anne
  Lachaux, Timoth{\'{e}}e Lacroix, Baptiste Rozi{\`{e}}re, Naman Goyal, Eric
  Hambro, Faisal Azhar, Aur{\'{e}}lien Rodriguez, Armand Joulin, Edouard Grave,
  and Guillaume Lample. 2023.
\newblock \href {https://doi.org/10.48550/arXiv.2302.13971} {Llama: Open and
  efficient foundation language models}.
\newblock \emph{CoRR}, abs/2302.13971.

\bibitem[{Turney(2002)}]{acl02-thumbs}
Peter~D. Turney. 2002.
\newblock \href {https://aclanthology.org/P02-1053.pdf} {Thumbs up or thumbs
  down? semantic orientation applied to unsupervised classification of
  reviews}.
\newblock In \emph{ACL}, pages 417--424.

\bibitem[{Varathan et~al.(2017)Varathan, Giachanou, and
  Crestani}]{comparative-survey}
Kasturi~Dewi Varathan, Anastasia Giachanou, and Fabio Crestani. 2017.
\newblock \href {https://doi.org/10.1002/asi.23716} {Comparative opinion
  mining: {A} review}.
\newblock \emph{J. Assoc. Inf. Sci. Technol.}, 68(4):811--829.

\bibitem[{Wang et~al.(2019)Wang, Singh, Michael, Hill, Levy, and Bowman}]{glue}
Alex Wang, Amanpreet Singh, Julian Michael, Felix Hill, Omer Levy, and
  Samuel~R. Bowman. 2019.
\newblock \href {https://openreview.net/forum?id=rJ4km2R5t7} {{GLUE:} {A}
  multi-task benchmark and analysis platform for natural language
  understanding}.
\newblock In \emph{7th International Conference on Learning Representations,
  {ICLR} 2019, New Orleans, LA, USA, May 6-9, 2019}. OpenReview.net.

\bibitem[{Wang et~al.(2023)Wang, Xie, Ding, Feng, and Xia}]{llm-sa-xiarui}
Zengzhi Wang, Qiming Xie, Zixiang Ding, Yi~Feng, and Rui Xia. 2023.
\newblock \href {https://doi.org/10.48550/arXiv.2304.04339} {Is chatgpt a good
  sentiment analyzer? {A} preliminary study}.
\newblock \emph{CoRR}, abs/2304.04339.

\bibitem[{Wei et~al.(2022)Wei, Bosma, Zhao, Guu, Yu, Lester, Du, Dai, and
  Le}]{instruction-tuning}
Jason Wei, Maarten Bosma, Vincent~Y. Zhao, Kelvin Guu, Adams~Wei Yu, Brian
  Lester, Nan Du, Andrew~M. Dai, and Quoc~V. Le. 2022.
\newblock \href {https://openreview.net/forum?id=gEZrGCozdqR} {Finetuned
  language models are zero-shot learners}.
\newblock In \emph{The Tenth International Conference on Learning
  Representations, {ICLR} 2022, Virtual Event, April 25-29, 2022}.

\bibitem[{Xu et~al.(2020)Xu, Li, Lu, and Bing}]{xu-etal-2020-aste}
Lu~Xu, Hao Li, Wei Lu, and Lidong Bing. 2020.
\newblock \href {https://aclanthology.org/2020.emnlp-main.183} {Position-aware
  tagging for aspect sentiment triplet extraction}.
\newblock In \emph{Proceedings of the 2020 Conference on Empirical Methods in
  Natural Language Processing (EMNLP)}, pages 2339--2349.

\bibitem[{Yadav and Vishwakarma(2020)}]{aireview20-sa-dl-survey}
Ashima Yadav and Dinesh~Kumar Vishwakarma. 2020.
\newblock \href {https://doi.org/10.1007/s10462-019-09794-5} {Sentiment
  analysis using deep learning architectures: a review}.
\newblock \emph{Artif. Intell. Rev.}, 53(6):4335--4385.

\bibitem[{Yang et~al.(2023)Yang, Jin, Tang, Han, Feng, Jiang, Yin, and
  Hu}]{chatgpt-survey}
Jingfeng Yang, Hongye Jin, Ruixiang Tang, Xiaotian Han, Qizhang Feng, Haoming
  Jiang, Bing Yin, and Xia Hu. 2023.
\newblock \href {https://doi.org/10.48550/arXiv.2304.13712} {Harnessing the
  power of llms in practice: {A} survey on chatgpt and beyond}.
\newblock \emph{CoRR}, abs/2304.13712.

\bibitem[{Ye et~al.(2023)Ye, Chen, Xu, Zu, Shao, Liu, Cui, Zhou, Gong, Shen,
  Zhou, Chen, Gui, Zhang, and Huang}]{gpt35-evaluate}
Junjie Ye, Xuanting Chen, Nuo Xu, Can Zu, Zekai Shao, Shichun Liu, Yuhan Cui,
  Zeyang Zhou, Chao Gong, Yang Shen, Jie Zhou, Siming Chen, Tao Gui, Qi~Zhang,
  and Xuanjing Huang. 2023.
\newblock \href {https://doi.org/10.48550/arXiv.2303.10420} {A comprehensive
  capability analysis of {GPT-3} and {GPT-3.5} series models}.
\newblock \emph{CoRR}, abs/2303.10420.

\bibitem[{Yu and Hatzivassiloglou(2003)}]{emnlp02-yu-sa}
Hong Yu and Vasileios Hatzivassiloglou. 2003.
\newblock \href {https://aclanthology.org/W03-1017/} {Towards answering opinion
  questions: Separating facts from opinions and identifying the polarity of
  opinion sentences}.
\newblock In \emph{Proceedings of the Conference on Empirical Methods in
  Natural Language Processing, {EMNLP} 2003, Sapporo, Japan, July 11-12, 2003}.

\bibitem[{Yue et~al.(2019)Yue, Chen, Li, Zuo, and Yin}]{sa-survey-scoial-media}
Lin Yue, Weitong Chen, Xue Li, Wanli Zuo, and Minghao Yin. 2019.
\newblock \href {https://doi.org/10.1007/s10115-018-1236-4} {A survey of
  sentiment analysis in social media}.
\newblock \emph{Knowl. Inf. Syst.}, 60(2):617--663.

\bibitem[{Zampieri et~al.(2019)Zampieri, Malmasi, Nakov, Rosenthal, Farra, and
  Kumar}]{offensive}
Marcos Zampieri, Shervin Malmasi, Preslav Nakov, Sara Rosenthal, Noura Farra,
  and Ritesh Kumar. 2019.
\newblock \href {https://doi.org/10.18653/v1/s19-2010} {Semeval-2019 task 6:
  Identifying and categorizing offensive language in social media
  (offenseval)}.
\newblock In \emph{Proceedings of the 13th International Workshop on Semantic
  Evaluation, SemEval@NAACL-HLT 2019, Minneapolis, MN, USA, June 6-7, 2019},
  pages 75--86. Association for Computational Linguistics.

\bibitem[{Zeng and Li(2022)}]{irony-survey}
Qingcheng Zeng and An{-}Ran Li. 2022.
\newblock \href {https://aclanthology.org/2022.coling-1.69} {A survey in
  automatic irony processing: Linguistic, cognitive, and multi-x perspectives}.
\newblock In \emph{Proceedings of the 29th International Conference on
  Computational Linguistics, {COLING} 2022, Gyeongju, Republic of Korea,
  October 12-17, 2022}, pages 824--836. International Committee on
  Computational Linguistics.

\bibitem[{Zhang et~al.(2021)Zhang, Deng, Li, Yuan, Bing, and Lam}]{asqp}
Wenxuan Zhang, Yang Deng, Xin Li, Yifei Yuan, Lidong Bing, and Wai Lam. 2021.
\newblock \href {https://aclanthology.org/2021.emnlp-main.726} {Aspect
  sentiment quad prediction as paraphrase generation}.
\newblock In \emph{Proceedings of the 2021 Conference on Empirical Methods in
  Natural Language Processing}, pages 9209--9219. Association for Computational
  Linguistics.

\bibitem[{Zhang et~al.(2022)Zhang, Li, Deng, Bing, and Lam}]{absa-survey}
Wenxuan Zhang, Xin Li, Yang Deng, Lidong Bing, and Wai Lam. 2022.
\newblock \href {https://doi.org/10.48550/arXiv.2203.01054} {A survey on
  aspect-based sentiment analysis: Tasks, methods, and challenges}.
\newblock \emph{CoRR}, abs/2203.01054.

\bibitem[{Zhang et~al.(2015)Zhang, Zhao, and LeCun}]{yelp}
Xiang Zhang, Junbo~Jake Zhao, and Yann LeCun. 2015.
\newblock \href
  {https://proceedings.neurips.cc/paper/2015/hash/250cf8b51c773f3f8dc8b4be867a9a02-Abstract.html}
  {Character-level convolutional networks for text classification}.
\newblock In \emph{Advances in Neural Information Processing Systems 28: Annual
  Conference on Neural Information Processing Systems 2015, December 7-12,
  2015, Montreal, Quebec, Canada}, pages 649--657.

\bibitem[{Zhong et~al.(2023)Zhong, Ding, Liu, Du, and Tao}]{llm-sa-chatgpt}
Qihuang Zhong, Liang Ding, Juhua Liu, Bo~Du, and Dacheng Tao. 2023.
\newblock \href {https://arxiv.org/abs/2302.10198} {Can chatgpt understand too?
  {A} comparative study on chatgpt and fine-tuned {BERT}}.
\newblock \emph{CoRR}, abs/2302.10198.

\end{thebibliography}
\bibliographystyle{acl_natbib}

\appendix

\section{Appendix}
\label{sec:appendix}

\subsection{Prompts for Each SA Task}
\label{prompt_summary}
We present a 1-shot prompt for each investigated sentiment analysis task, which is shown on the following pages.

\onecolumn 
\footnotesize{
\begin{longtable}{|c|c|p{0.7\linewidth}|}
\hline
\textbf{task} & \textbf{Dataset} & \multicolumn{1}{c}{\textbf{1-shot Prompt}} \\
\hline
\endfirsthead 
\hline
\multicolumn{3}{|c|}{\textit{Continued from previous page}} \\
\hline
\endhead 
\hline
\multicolumn{3}{|c|}{\textit{Continued on next page}} \\
\hline
\endfoot 
\endlastfoot 
SC & IMDb & Please perform Sentiment Classification task.
Given the sentence, assign a sentiment label from ['negative', 'positive'].
Return label only without any other text.
\newline

Sentence:
I 've seen the original English version on video . Disney 's choice of voice actors looks very promising ....

Label:positive

Sentence:
`` This is a depressingly shallow , naive and mostly unfunny look at a wildly improbable relationship between Brooks ' psychotic film editor and Harold , his vapid girlfriend ....

Label:negative
\newline

Sentence:
`` Jack and Kate meet the physician Daniel Farady first and then the psychics Miles Straume and they demonstrate that have not come to the island with the intention of rescuing the survivors . Locke and his group find the anthropologist Charlotte Staples Lewis , and Ben Linus shoots her . Meanwhile , the group of Jack finds the pilot Frank Lapidus , who landed the helicopter with minor damages that can be repaired . Jack forces Miles to tell the real intention why they have come to the island. < br / > < br / > The second episode of the Fourth Season returns to the island , with four new characters , stops the confusing `` '' flash-forwards '' '' and it seems that will finally be the beginning of the explanations that I ( and most of the fans and viewers ) expect to be provided in `` '' Lost '' '' . Why the interest of the government in Ben Linus , and how he is informed from the boat are some of the questions that I expect to see in the next episodes . My vote is eight. < br / > < br / > Title ( Brazil ) : Not Available ''

Label: \\
\hline
SC & Yelp-2 & Please perform Sentiment Classification task.
Given the sentence, assign a sentiment label from ['negative', 'positive'].
Return label only without any other text.
\newline

Sentence:
Had a great time with my beautiful wife listening to The Instant Classics . Drinks are pricey and menu seems a little limited , but I had a great time .... 

Label:positive

Sentence:
I have been to this location multiple times and every time the service is horrendous and the food is mediocre . Not sure if the location being in a mall has to do with it ....

Label:negative
\newline

Sentence:
I expected the prices of the entrees to be a little bit higher but the quality of the Chinese food was not worth the money I paid for the dishes . I got the 18 monk noodle and the traditional dimsum . If I could describe the food in one word-terrible ! Making the dimsum look pretty by topping it with gold flakes did not do anything to make up for the flavor of the dimsum . It seemed too starchy and you can hardly taste the meat . The noodles looked like a sad , greasy slop of Mai fun type noodles ( noodles were stuck together ) saturated with soy sauce for color , and garnished with a few pieces of shitake mushrooms , green onions and fine threads of carrots . And yes , portions were small , but that 's not really the worst part of the whole experience . Just poorly prepared , way overpriced Chinese food ... sorry .

Label:\\
\hline
SC & Yelp-5 & Please perform Sentiment Classification task.
Given the sentence, assign a sentiment label from ['negative', 'neutral', 'positive', 'very negative', 'very positive'].
Return label only without any other text.
\newline

Sentence:
The most important thing to me in an airline is that we do not fall out of the sky in an uncontrolled fashion . After all landing is a controlled crash ....

Label:neutral

Sentence:
`` Great place to go for hair , nails or massage . Great service in a professional and clean environment . Most places u have to wait even if u have an appt ....

Label:very positive

Sentence:
Loved the atmosphere . Right across from chase field . The pretzel and provolone and shrimp appetizers were plentiful and fantastic . Easily enough for four people to share ....

Label:positive

Sentence:
`` 1 star- why ? The food was n't too bad . My husband had the fish tacos which were good . I ordered the Sicilian Stuffed Chicken , but get this .... 

Label:negative

Sentence:
`` Hello there ! 00a0 00a0 00a0 My name is Naiby Moreno , and the reason why I 'm writing you this email is because last night , around this time ....

Label:very negative
\newline

Sentence:
Came a few days ago for a lease , was n't sure of size needed , so I guessed , three times ! Finally got it right , but hey , the store did n't bat a eye lash when I returned the ones that did n't work , they just asked if I needed help picking out a replacement . Since my cat has been loosing weight , I could not get the size down , so after my attempts , finally got the small dog size and sure enough it worked . Now to get the cat used to it before we need it . This store has everything you could need . They is even a new section by Martha Stewart , everything for you little pet . But her stuffs pricey , a lease from here collection , \$ 19.99 , boy that 's steep ! The store is clean , neatly kept , well organized and they have grooming services . The employees were friendly and helpful , they looked like they enjoyed their jobs , and I would make this a regular place .

Label: \\
\hline
SC & MR & Please perform Sentiment Classification task.
Given the sentence, assign a sentiment label from ['negative', 'positive'].
Return label only without any other text.
\newline

Sentence:
`` it 's the chemistry between the women and the droll scene-stealing wit and wolfish pessimism of anna chancellor that makes this `` '' two weddings and a funeral `` '' fun . ''

Label:positive

Sentence:
the entire movie is about a boring , sad man being boring and sad .

Label:negative
\newline

Sentence:
`` if you 're a crocodile hunter fan , you 'll enjoy at least the `` '' real `` '' portions of the film . if you 're looking for a story , do n't bother . ''

Label: \\
\hline 
SC & SST2 & Please perform Sentiment Classification task.
Given the sentence, assign a sentiment label from ['negative', 'positive'].
Return label only without any other text.
\newline

Sentence:
Oh , and more entertaining , too .

Label:positive

Sentence:
If you 're not a fan , it might be like trying to eat Brussels sprouts .

Label:negative
\newline

Sentence:
An ungainly , comedy-deficient , B-movie rush job ...

Label:\\
\hline
SC & Twitter & Please perform Sentiment Classification task.
Given the sentence, assign a sentiment label from ['negative', 'neutral', 'positive'].
Return label only without any other text.
\newline

Sentence:
- Just bought my 1st iPad, iPad3, feeling real burned, mad, about iPad4 so soon. Grrr. REALLY mad! Don't even care about mini now,"

Label:negative

Sentence:
@user @user @user  I think this is the motive of the Yakub's laywers for pursuing the case

Label:neutral

Sentence:
Kanye West was honored in a big way during Sunday night's MTV Video Music Awards by receiving the Michael Jackso...

Label:positive
\newline

Sentence:
Do you think Michelle Obama wanted to smack Melania Trump for plagiarizing her convention speech? She has the arms for it.

Label: \\
\hline
SC & SST5 & Please perform Sentiment Classification task.
Given the sentence, assign a sentiment label from ['negative', 'neutral', 'positive', 'very negative', 'very positive'].
Return label only without any other text.
\newline

Sentence:
` Like a child with an important message to tell ... ( Skins ' ) faults are easy to forgive because the intentions are lofty . '

Label:neutral

Sentence:
That Haynes can both maintain and dismantle the facades that his genre and his character construct is a wonderous accomplishment of veracity and narrative grace .

Label:very positive

Sentence:
Oh , and more entertaining , too .

Label:positive

Sentence:
If you 're not a fan , it might be like trying to eat Brussels sprouts .

Label:negative

Sentence:
When it comes out on video , then it 's the perfect cure for insomnia .

Label:very negative
\newline

Sentence:
Everywhere the camera looks there is something worth seeing .

Label:\\
\hline
SC & Lap14 & Please perform Aspect Sentiment Classification task.
Given the sentence, assign a sentiment label towards "Office" from ['negative', 'neutral', 'positive'].
Return label only without any other text.
\newline

Sentence:
It even has a great webcam , and Skype works very well . (sentiment towards "webcam")

Label:positive

Sentence:
- Touchpad will take a bit of time to get used to . (sentiment towards "- Touchpad")

Label:neutral

Sentence:
) And printing from either word processor is an adventure . (sentiment towards "word processor")

Label:negative
\newline

Sentence:
( but Office can be purchased ) IF I ever need a laptop again I am for sure purchasing another Toshiba !!

Label:
\\
\hline
SC & Rest14 & Please perform Aspect Sentiment Classification task.
Given the sentence, assign a sentiment label towards "garlic knots" from ['negative', 'neutral', 'positive'].
Return label only without any other text.
\newline

Sentence:
While the new restaurant still features much of the same classical furniture that made Tiffin so attractive , the menu has been overhauled . (sentiment towards "classical furniture")

Label:positive

Sentence:
And it all comes at a very reasonable price ( congee , noodles , and rice dishes are no more than  3-6 each ) . (sentiment towards "( congee")

Label:neutral

Sentence:
The Singapore Mai Fun had NO curry flavor whatsoever . (sentiment towards "curry flavor")

Label:negative
\newline

Sentence:
I also recommend the garlic knots .

Label:
\\
\hline
UABSA & Rest14 & Please perform Unified Aspect-Based Sentiment Analysis task.
Given the sentence, tag all (aspect, sentiment) pairs. Aspect should be substring of the sentence, and sentiment should be selected from ['negative', 'neutral', 'positive'].
If there are no aspect-sentiment pairs, return an empty list. Otherwise return a python list of tuples containing two strings in double quotes. Please return python list only, without any other comments or texts.
\newline

Sentence:
also make sure you pay attention to the music being piped in , quite a weird selection .

Label:[('music', 'neutral')]

Sentence:
but I would n't wan na live there .

Label:[]

Sentence:
And their prices are very high , they actually think that they can get away with charging such prices for such terrible food and service !

Label:[('prices', 'negative'), ('prices', 'negative'), ('food', 'negative'), ('service', 'negative')]

Sentence:
Having not been home in the last 2 years may skew this reviewer a bit , but the food was tasty and spicy sans the oil that comes floating along at similar venues .

Label:[('food', 'positive'), ('oil', 'neutral')]
\newline

Sentence:
After I paid for my purchase , I noticed they had not given me utensils so I could eat my pie .

Label:\\
\hline
UABSA & Rest15 &
Please perform Unified Aspect-Based Sentiment Analysis task.
Given the sentence, tag all (aspect, sentiment) pairs. Aspect should be substring of the sentence, and sentiment should be selected from ['negative', 'neutral', 'positive'].
If there are no aspect-sentiment pairs, return an empty list. Otherwise return a python list of tuples containing two strings in double quotes. Please return python list only, without any other comments or texts.
\newline

Sentence:
The portions are HUGE , so it might be good to order three things to split rather than one appetizer and entree per person for two people .

Label:[('portions', 'neutral')]

Sentence:
No , really .

Label:[]

Sentence:
The food was bland oily .

Label:[('food', 'negative')]

Sentence:
The food 's as good as ever .

Label:[('food', 'positive')]
\newline

Sentence:
Need I say more ?

Label: \\
\hline
UABSA & Rest16 & Please perform Unified Aspect-Based Sentiment Analysis task.
Given the sentence, tag all (aspect, sentiment) pairs. Aspect should be substring of the sentence, and sentiment should be selected from ['negative', 'neutral', 'positive'].
If there are no aspect-sentiment pairs, return an empty list. Otherwise return a python list of tuples containing two strings in double quotes. Please return python list only, without any other comments or texts.
\newline

Sentence:
Food was okay , nothing great .

Label:[('Food', 'neutral')]

Sentence:
I live in the neightborhood and am a regular .

Label:[]

Sentence:
The place is small and cramped but the food is fantastic .

Label:[('place', 'negative'), ('food', 'positive')]

Sentence:
One special roll and one regular roll is enough to fill you up , but save room for dessert !

Label:[('special roll', 'positive'), ('regular roll', 'positive'), ('dessert', 'positive')]
\newline

Sentence:
The atmosphere is aspiring , and the decor is festive and amazing .

Label: \\
\hline
UABSA & Laptop14 & Please perform Unified Aspect-Based Sentiment Analysis task.
Given the sentence, tag all (aspect, sentiment) pairs. Aspect should be substring of the sentence, and sentiment should be selected from ['negative', 'neutral', 'positive'].
If there are no aspect-sentiment pairs, return an empty list. Otherwise return a python list of tuples containing two strings in double quotes. Please return python list only, without any other comments or texts.
\newline

Sentence:
After that the said it was under warranty .

Label:[('warranty', 'neutral')]

Sentence:
I really wanted a Mac over a pc because I used a Mac in high school .

Label:[]

Sentence:
Another issue I have with it is the battery .

Label:[('battery', 'negative')]

Sentence:
I love the size , keyboard , the functions .

Label:[('size', 'positive'), ('keyboard', 'positive'), ('functions', 'positive')]
\newline

Sentence:
Hopefully my replacement is brand new .

Label: \\
\hline
ASTE & Rest 14 & Please perform Aspect Sentiment Triplet Extraction task.
Given the sentence, tag all (aspect, opinion, sentiment) triplets. Aspect and opinion should be substring of the sentence, and sentiment should be selected from ['negative', 'neutral', 'positive'].
Return a python list of tuples containing three strings in double quotes. Please return python list only, without any other comments or texts.
\newline

Sentence:
Service was slow had to wait to order and get food although not crowded .

Label:[('Service', 'slow', 'negative')]

Sentence:
The atmosphere is n't the greatest , but I suppose that 's how they keep the prices down .

Label:[('atmosphere', "is n't the greatest", 'neutral'), ('prices', 'down', 'positive')]

Sentence:
The fries are yummy .

Label:[('fries', 'yummy', 'positive')]
\newline

Sentence:
Most importantly , it is reasonably priced .

Label:
\\
\hline
ASTE & Rest 15 & Please perform Aspect Sentiment Triplet Extraction task.
Given the sentence, tag all (aspect, opinion, sentiment) triplets. Aspect and opinion should be substring of the sentence, and sentiment should be selected from ['negative', 'neutral', 'positive'].
Return a python list of tuples containing three strings in double quotes. Please return python list only, without any other comments or texts.
\newline

Sentence:
the only things u could really taste are the very salty soy sauce ( even its low sodium ) , the vinegar-soaked rice , and the scallion on top of the fish .

Label:[('soy sauce', 'salty', 'negative'), ('rice', 'vinegar-soaked', 'negative')]

Sentence:
Food was okay , nothing great .

Label:[('Food', 'okay', 'neutral'), ('Food', 'nothing great', 'neutral')]

Sentence:
We recently decided to try this location , and to our delight , they have outdoor seating , perfect since I had my yorkie with me .

Label:[('outdoor seating', 'perfect', 'positive')]
\newline

Sentence:
This establishment is the real deal .

Label:
\\
\hline
ASTE & Rest 16 & Please perform Aspect Sentiment Triplet Extraction task.
Given the sentence, tag all (aspect, opinion, sentiment) triplets. Aspect and opinion should be substring of the sentence, and sentiment should be selected from ['negative', 'neutral', 'positive'].
Return a python list of tuples containing three strings in double quotes. Please return python list only, without any other comments or texts.
\newline

Sentence:
limited menu , no-so-fresh ingredients , thinly-sliced fish , fall-apart rice .

Label:[('menu', 'limited', 'negative'), ('ingredients', 'no-so-fresh', 'negative'), ('fish', 'thinly-sliced', 'negative'), ('rice', 'fall-apart', 'negative')]

Sentence:
For desserts , we tried the frozen black sesame mousse ( interesting but not extraordinary ) and matcha ( powdered green tea ) and blueberry cheesecake , which was phenomenal .

Label:[('frozen black sesame mousse', 'interesting', 'neutral'), ('frozen black sesame mousse', 'extraordinary', 'neutral'), ('matcha ( powdered green tea ) and blueberry cheesecake', 'phenomenal', 'positive')]

Sentence:
The food was good .

Label:[('food', 'good', 'positive')]
\newline

Sentence:
In Grammercy/Union Square/East Village this is my neighbors and my favorite spot .

Label:
\\
\hline
ASTE & Laptap14 & Please perform Aspect Sentiment Triplet Extraction task.
Given the sentence, tag all (aspect, opinion, sentiment) triplets. Aspect and opinion should be substring of the sentence, and sentiment should be selected from ['negative', 'neutral', 'positive'].
Return a python list of tuples containing three strings in double quotes. Please return python list only, without any other comments or texts.
\newline

Sentence:
Dealing with the support drone on the other end of the chat was sheer torture .

Label:[('support', 'sheer torture', 'negative')]

Sentence:
I did think it had a camera because that was one of my requirements , but forgot to check in the specifications on this one before I purchased .

Label:[('specifications', 'check in', 'neutral')]

Sentence:
A longer battery life would have been great - but it meets it 's spec quite easily .

Label:[('spec', 'easily', 'positive')]
\newline

Sentence:
It was important that it was powerful enough to do all of the tasks he needed on the internet , word processing , graphic design and gaming .

Label:
\\
\hline
ASQP & Rest15 & Please perform Aspect Sentiment Quad Prediction task.
Given the sentence, tag all (category, aspect, opinion, sentiment) quadruples. Aspect and opinion should be substring of the sentence. Category should be selected from ['ambience general', 'drinks prices', 'drinks quality', 'drinks style\_options', 'food general', 'food prices', 'food quality', 'food style\_options', 'location general', 'restaurant general', 'restaurant miscellaneous', 'restaurant prices', 'service general']. Sentiment should be selected from ['negative', 'neutral', 'positive']. Only aspect can be 'NULL', category, opinion and sentiment cannot be 'NULL'.
Return a python list of tuples containing four strings in double quotes. Please return python list only, without any other comments or texts.
\newline

Sentence:
The price is reasonable although the service is poor .

Label:[('restaurant prices', 'NULL', 'reasonable', 'positive'), ('service general', 'service', 'poor', 'negative')]

Sentence:
This little place definitely exceeded my expectations and you sure get a lot of food for your money .

Label:[('food style\_options', 'food', 'lot', 'positive'), ('restaurant general', 'place', 'exceeded my expectations', 'positive'), ('food prices', 'food', 'lot', 'positive')]

Sentence:
This place is really trendi but they have forgotten about the most important part of a restaurant , the food .

Label:[('food quality', 'food', 'forgotten', 'negative'), ('ambience general', 'place', 'trendi', 'positive')]

Sentence:
The restaurant looks out over beautiful green lawns to the Hudson River and the Statue of Liberty .

Label:[('location general', 'restaurant', 'beautiful', 'positive')]

Sentence:
With so many good restaurants on the UWS , I do n't need overpriced food , absurdly arrogant wait-staff who do n't recognize they work at a glorified diner , clumsy service , and management that does n't care .

Label:[('food prices', 'food', 'overpriced', 'negative'), ('service general', 'wait-staff', 'arrogant', 'negative'), ('service general', 'service', 'clumsy', 'negative'), ('service general', 'management', "does n't care", 'negative')]

Sentence:
the drinks are amazing and half off till 8pm .

Label:[('drinks quality', 'drinks', 'amazing', 'positive'), ('drinks prices', 'drinks', 'amazing', 'positive')]

Sentence:
A cool bar with great food , and tons of excellent beer .

Label:[('ambience general', 'bar', 'cool', 'positive'), ('food quality', 'food', 'great', 'positive'), ('drinks quality', 'beer', 'excellent', 'positive'), ('drinks style\_options', 'beer', 'excellent', 'positive')]

Sentence:
The food is great and reasonably priced .

Label:[('food quality', 'food', 'great', 'positive'), ('food prices', 'food', 'reasonably priced', 'positive')]
....
\newline

Sentence:
For me dishes a little oily , but overall dining experience good .

Label:
\\
\hline
ASQP & Rest16 & Please perform Aspect Sentiment Quad Prediction task.
Given the sentence, tag all (category, aspect, opinion, sentiment) quadruples. Aspect and opinion should be substring of the sentence. Category should be selected from ['ambience general', 'drinks prices', 'drinks quality', 'drinks style\_options', 'food general', 'food prices', 'food quality', 'food style\_options', 'location general', 'restaurant general', 'restaurant miscellaneous', 'restaurant prices', 'service general']. Sentiment should be selected from ['negative', 'neutral', 'positive']. Only aspect can be 'NULL', category, opinion and sentiment cannot be 'NULL'.
Return a python list of tuples containing four strings in double quotes. Please return python list only, without any other comments or texts.
\newline

Sentence:
The wine list is interesting and has many good values .

Label:[('drinks style\_options', 'wine list', 'interesting', 'positive'), ('drinks prices', 'wine list', 'good values', 'positive')]

Sentence:
The food is amazing ... especially if you get the Chef 's tasting menu and your favourite bottle ( or two ! ) of wine from an extensive selection of wines .
k

Label:[('food quality', 'food', 'amazing', 'positive'), ('drinks style\_options', 'selection of wines', 'extensive', 'positive'), ('food quality', "Chef 's tasting menu", 'favourite', 'positive')]

Sentence:
Gorgeous place ideal for a romantic dinner

Label:[('ambience general', 'place', 'Gorgeous', 'positive'), ('restaurant miscellaneous', 'place', 'ideal', 'positive')]

Sentence:
The drinks are great , especially when made by Raymond .

Label:[('drinks quality', 'drinks', 'great', 'positive'), ('service general', 'Raymond', 'great', 'positive')]....
\newline

Sentence:
It was worth the wait .

Label:

\\
\hline
Implicit & Lap+Res & Please perform Aspect-Based Implicit Sentiment Analysis task.
Given the sentence, please infer the sentiment towards the aspect "vintages". Please select a sentiment label from ['negative', 'neutral', 'positive'].
Return label only without any other text.
\newline

Sentence:
The steak was excellent and one of the best I have had (I tasted the butter intitally but in no way did it overwhelm the flavor of the meat). (sentiment towards "butter")

Label:negative

Sentence:
Yes, they use fancy ingredients, but even fancy ingredients don't make for good pizza unless someone knows how to get the crust right. (sentiment towards "crust")

Label:neutral

Sentence:
Three page wine menu, one page entree and horedevous. (sentiment towards "wine menu")

Label:positive
\newline

Sentence:
Somewhat disappointing wine list (only new vintages.

Label:
\\
\hline
Hate & HatEval & 
Please perform Hate Detection task.
Given the sentence, assign a sentiment label from ['hate', 'non-hate'].
Return label only without any other text.
\newline

Sentence:
My family’s idea of a merienda for this moment is siopao. They really hate me. Me: *calls Tim Ho Wan* Do you deliver in elyu? 

Label:non-hate

Sentence:
This is horrendous

Label:hate
\newline

Sentence:
@user id marry this fukin whore, let the bitch behind her be best lady at the wedding

Label:

\\
\hline
Irony & Irony18 & Please perform Irony Detection task.
Given the sentence, please determine wheter or not it contains irony. Assign a sentiment label from ['irony', 'non\_irony'].
Return label only without any other text.
\newline

Sentence:
@user You truly are my son.

Label:non\_irony

Sentence:
Just watched how Pretzels were made. 

Label:irony
\newline

Sentence:
Fighting over chargers is definitely how I wanted to start my day.

Label:
\\
\hline
Offensive & OffensEval & Please perform Offensive Detection task.
Given the sentence, assign a sentiment label from ['non-offensive', 'offensive'].
Return label only without any other text.
\newline

Sentence:
user  Hi Bernice  I hope you are enjoying the xrpcommunity and learning lots about xrp  0589 user 

Label:non-offensive

Sentence:
@user this isn't me disagreeing this is me basically saying that i hope you're right but if you are i will spontaneously combust

Label:offensive
\newline

Sentence:
MAGA  ... got any ideas how she could have done it?

Label:
\\
\hline
Stance & Stance16 & Please perform Stance Detection (abortion) task.
Given the sentence, assign a sentiment label expressed by the author towards "abortion" from ['against', 'favor', 'none'].
Return label only without any other text.
\newline

Sentence:
\@user i don't follow the news, is there a new law that ALL gay people have to get married? I'm against that! \#SemST (opinion towards "abortion")

Label:none

Sentence:
The natural world is part of our inheritance, we have to protect it  \@user with \@user on \#BBC \#Earth \#SemST (opinion towards "climate")

Label:favor

Sentence:
\@user we lost 4,000 of our Military boys when your President pulled out of Iraq.  \#LiberalConsequences  \#SemST (opinion towards "hillary")

Label:against
\newline

Sentence:
Women have outgrown the common housewife stigma long ago \#SemST

Label:
\\
\hline
Comparative & CS19 & Please perform Comparative Opinions task.
Given the sentence, compare "Microsoft" to "Sony", and assign an opinion label from ['better', 'worse'].
Return label only without any other text.
\newline

Sentence:
Java isn't too bad of a first language, but Python is a little easier to pick up. (compare "Java" to "Python")

Label:worse

Sentence:
In supply-chain conversations, the Pacific Crest semiconductor team learned that Windows 7 inventory is moving faster than Windows 8. (compare "Windows 7" to "Windows 8")

Label:better
\newline

Sentence:
And I think Microsoft will have more money to make better games than Sony.

Label:
\\
\hline
Emotion & Emotion20 & Please perform Comparative Opinions task.
Given the sentence, compare "Microsoft" to "Sony", and assign an opinion label from ['better', 'worse'].
Return label only without any other text.
\newline

Sentence:
the football team is decent but getting better! the basketball teams are awesome!the

Label:worse

Sentence:
Now let's be clear; in this author's humble opinion, Apple is still way better than IBM.

Label:better
\newline

Sentence:
And I think Microsoft will have more money to make better games than Sony.

Label: 
\\
\hline

\caption{Detailed prompts for investigated tasks and datasets. We show 1-shot prompt for illustration.} 
\end{longtable}
}
\twocolumn

\end{document}